\documentclass[runningheads]{llncs}
\usepackage{graphicx}
\usepackage{amsmath,amssymb} 
\usepackage{tabularx}
\usepackage{multirow}
\usepackage{subfigure}
\usepackage{dsfont}
\usepackage{colortbl}
\usepackage[table]{xcolor}
\usepackage{color}\usepackage[width=122mm,left=12mm,paperwidth=146mm,height=193mm,top=12mm,paperheight=217mm]{geometry}

\newcommand{\figref}[1]{Fig. \ref{#1}}
\newcommand{\tabref}[1]{Table \ref{#1}}
\newcommand{\equref}[1]{(\ref{#1})}
\newcommand{\secref}[1]{Sec. \ref{#1}}

\makeatletter
\def\hlinewd#1{%
	\noalign{\ifnum0=`}\fi\hrule \@height #1 \futurelet
	\reserved@a\@xhline}

\begin{document}
	\pagestyle{headings}
	\mainmatter
	\title{Unified Depth Prediction and Intrinsic Image Decomposition from a Single Image\\ via Joint Convolutional Neural Fields} 
	
	\titlerunning{Unified Depth Prediction and Intrinsic Image Decomposition via JCNF}
	
	\authorrunning{S. Kim et al.}
	
	\author{Seungryong Kim$^{1}$\thanks{This work is done while Seungryong Kim was an intern at Microsoft Research.}, Kihong Park$^1$, Kwanghoon Sohn$^1$, and Stephen Lin$^2$}
	\institute{$^1$Yonsei University, $^2$Microsoft Research} 
	
	\maketitle
	\begin{abstract}
		We present a method for jointly predicting a depth map and intrinsic images from single-image input. The two tasks are formulated in a synergistic manner through a joint conditional random field (CRF) that is solved using a novel convolutional neural network (CNN) architecture, called the joint convolutional neural field (JCNF) model. Tailored to our joint estimation problem, JCNF differs from previous CNNs in its sharing of convolutional activations and layers between networks for each task, its inference in the gradient domain where there exists greater correlation between depth and intrinsic images, and the incorporation of a gradient scale network that learns the confidence of estimated gradients in order to effectively balance them in the solution. This approach is shown to surpass state-of-the-art methods both on single-image depth estimation and on intrinsic image decomposition.
		
		\keywords{single-image depth estimation, intrinsic image decomposition, conditional random field, convolutional neural networks}
	\end{abstract}

	\section{Introduction}\label{sec:1}
	Perceiving the physical properties of a scene undoubtedly plays a fundamental role
	in understanding real-world imagery. Such inherent properties include the
	3-D geometric configuration, the illumination or shading,
	and the reflectance or albedo of each scene surface.
	Depth prediction and intrinsic image decomposition, which aims to recover shading and albedo,
	are thus two fundamental yet challenging tasks in computer vision.
	While they address different aspects of scene understanding,
	there exist strong consistencies among depth and intrinsic images,
	such that information about one provides valuable prior knowledge for recovering
	the other.
	
	In the intrinsic image decomposition literature, several works have exploited
	measured depth information to make the decomposition problem more tractable
	\cite{Chen13,Laffont12,Lee12,Jeon14,Barron13}. These techniques have all
	demonstrated better performance than using RGB images alone.
	On the other hand, in the literature for single-image depth prediction,
	illumination-invariant features have been utilized for greater robustness
	in depth inference \cite{Eigen14,Fayao15}, and shading discontinuities have been used to detect surface boundaries \cite{Kong15}, suggesting that intrinsic images
	can be employed to enhance depth prediction performance.
	Although the two tasks are mutually beneficial, previous research have solved
	for them only in sequence, by using estimated intrinsic images to constrain
	depth prediction \cite{Kong15}, or vice versa \cite{Shelhamer15}.
	We propose in this paper to instead jointly predict depth and intrinsic images
	in a manner where the two complementary tasks can assist each other.
	
	We address this joint prediction problem using convolutional neural networks
	(CNNs), which have yielded state-of-the-art performance for the individual
	problems of single-image depth prediction \cite{Eigen14,Fayao15} and
	intrinsic image decomposition \cite{Shelhamer15,Zhou15,Narihira15}, but
	are hampered by ambiguity issues that arise from limited training sets.
	In our work, the two tasks are formulated synergistically in a joint conditional
	random field (CRF) that is solved using a novel CNN architecture,
	called the joint convolutional neural field (JCNF) model.
	This architecture differs from previous CNNs in several ways tailored to our
	particular problem. One is the sharing of convolutional activations and layers between
	networks for each task, which allows each network to account for inferences
	made in other networks. Another is to perform learning in the gradient
	domain, where there exist stronger correlations between depth and intrinsic
	images than in the image value domain, which helps to deal with
	the ambiguity problem from limited training sets.
	A third is the incorporation of a gradient scale network which jointly learns the confidence of the estimated gradients, to more robustly balance them in the solution.
	These networks of the JCNF model are jointly learned
	using a unified energy function in a joint CRF.
	
	Within this system, depth, shading and albedo are predicted in a coarse-to-fine
	manner that yields more globally consistent results. Our experiments show that
	this joint prediction outperforms existing depth prediction methods and
	intrinsic image decomposition techniques on various benchmarks.\vspace{-5pt}
	
	\section{Related Work}\label{sec:2}
	\vspace{-5pt}
	\subsubsection{Depth Prediction from a Single Image}\label{sec:21}
	Traditional methods for this task have formulated the depth prediction as a Markov random field (MRF) learning problem \cite{Saxena09,Wang14,Xiu14}. As exact MRF learning and inference are intractable in general, most of these approaches employ approximation methods, such as through linear regression of depth with image features \cite{Saxena09}, learning image-depth correlation with a non-linear kernel function \cite{Wang14}, and training category-adaptive model parameters \cite{Xiu14}. Although these parametric models infer plausible depth maps to some extent, they cannot estimate the depth of natural scenes reliably due to their limited learning capability.
	
	By leveraging the availability of large RGB-D databases,
	data-driven approaches have been actively researched \cite{Konrad13,Karsch14}.
	Konrad \emph{et al.} \cite{Konrad13} proposed a depth fusion scheme to infer the depth map
	by retrieving the nearest images in the dataset, followed by an aggregation via weighted median filtering.
	Karsch \emph{et al.} \cite{Karsch14} presented the depth transfer (DT) approach which retrieves the nearest similar images
	and warps their depth maps using dense SIFT flow.
	Inspired by this method,
	Choi \emph{et al.} \cite{Choi15} proposed the depth analogy (DA) approach
	that transfers depth gradients from the nearest images,
	demonstrating the effectiveness of gradient domain learning.
	Although these methods can extract reliable depth for certain scenes,
	there exist many others for which the nearest images are dissimilar and unsuitable. Recently, Kong \emph{et al.} \cite{Kong15} extended the DT approach \cite{Karsch14} by using albedo and shading for image matching as well as for
	detecting contours at surface boundaries. In contrast to our approach,
	the intrinsic images are estimated independently from the depth prediction.
	
	More recently, methods have been proposed based on CNNs. Eigen \emph{et al.} \cite{Eigen14} proposed multi-scale CNNs (MS-CNNs) for predicting depth maps directly from a
	single image. Other CNN models were later proposed for depth estimation
	\cite{Wang15}, including a deep convolutional neural field (DCNF) by
	Fayao \emph{et al.} \cite{Fayao15} that estimates depth on each superpixel
	while enforcing smoothness within a CRF.
	CNN-based methods clearly outperform conventional techniques, and
	we aim to elevate the performance further by accounting for intrinsic
	image information.
	\vspace{-8pt}
	
	\subsubsection{Intrinsic Image Decomposition}\label{sec:22}
	The notion of intrinsic images was first introduced in \cite{Barrow78}. Conventional methods are largely based on Retinex theory \cite{Land71,Shen08,Zhao12}, which attributes large image gradients to albedo changes, and smaller gradients to shading. More recent approaches have employed a variety of techniques, based on gradient distribution priors \cite{Li14}, dense CRFs \cite{Bell14}, and hybrid $L_2$-$L_p$ optimization to separate albedo and shading gradients \cite{Bonneel14}. These single-image based methods, however, are inherently limited by the fundamental ill-posedness of the problem. To partially alleviate this limitation, several approaches have utilized additional input,
	such as multiple images \cite{Wiess01,Laffont13,Kong14}, user interaction \cite{Bousseau09,Shen11}, and measured depth maps \cite{Chen13,Laffont12,Lee12,Jeon14,Barron13}. The use of additional data such as measured depth clearly increases performance but reduces their applicability.
	
	Related to our work is the method of Barron and Malik \cite{Barron12},
	which estimates object shape in addition to intrinsic images.
	To regularize the estimation, the method utilizes statistical priors
	on object shape and albedo which are not generally applicable to images
	of full scenes.
	
	More recently, intrinsic image decomposition has been addressed using CNNs \cite{Shelhamer15,Zhou15,Narihira15}.
	Zhou \emph{et al.} \cite{Zhou15} proposed a multi-stream CNN to predict the relative reflectance ordering between image patches from large-scale human annotations.
	Narihira \emph{et al.} \cite{Narihira15} learned a CNN that directly predicts albedo and shading from an RGB image patch.
	Shelhamer \emph{et al.} \cite{Shelhamer15} estimated depth through a fully convolutional network and used it to constrain the intrinsic image decomposition.
	Unlike our approach, the depth and intrinsic images are estimated sequentially. \vspace{-5pt}
	
	\section{Formulation}\label{sec:3}
	\vspace{-5pt}
	\subsection{Problem Statement and Model Architecture}\label{sec:31}
	Let us define a color image $I$ such that $I_{p}:\mathcal{I} \to {\mathbb R}^3$ for pixel $p$, where
	$\mathcal{I} \subset {{\mathbb N}^2}$ is a discrete image domain.
	Similarly, depth, albedo and shading can be defined as $D_{p}:\mathcal{I} \to {\mathbb R}$ and $A_{p},S_{p}:\mathcal{I} \to {\mathbb R}^3$. All images are defined in the log domain.
	Given a training set of color, depth, albedo, and shading images denoted by
	$\mathcal{C} = \{ \left({I^i},{D^i},{A^i},{S^i}\right)|i = 1,2,...,\mathcal{N}_\mathcal{C}\}$,
	where $\mathcal{N}_\mathcal{C}$ is the number of training images,
	we first aim to learn a prediction model that approximates
	depth $D^i$, albedo $A^i$, and shading $S^i$
	from each color image $I^i \in \mathcal{C}$.
	This prediction model will then be used to
	infer reliable depth $D$, albedo $A$, and shading $S$ simultaneously from a single query image $I$.
	
\begin{figure}[t]
		\centering
		\renewcommand{\thesubfigure}{}
		\subfigure[(]
		{\includegraphics[width=0.247\linewidth]{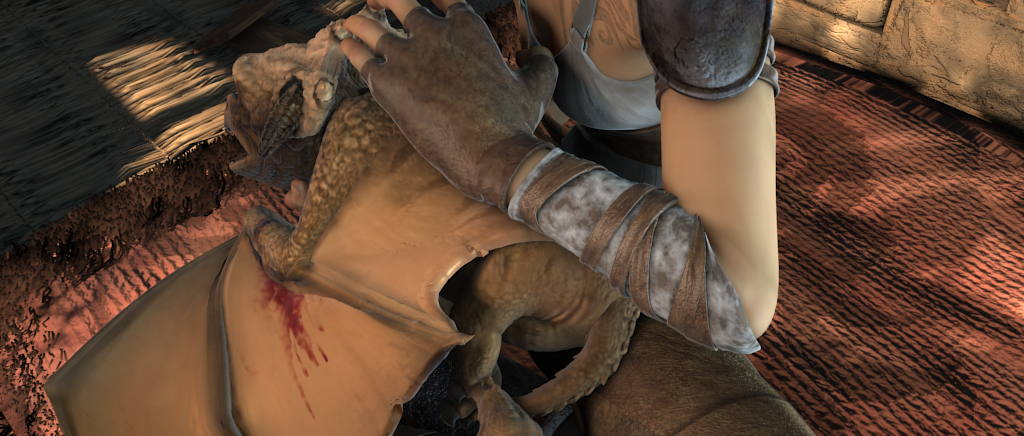}}\hfill
		\subfigure[]
		{\includegraphics[width=0.247\linewidth]{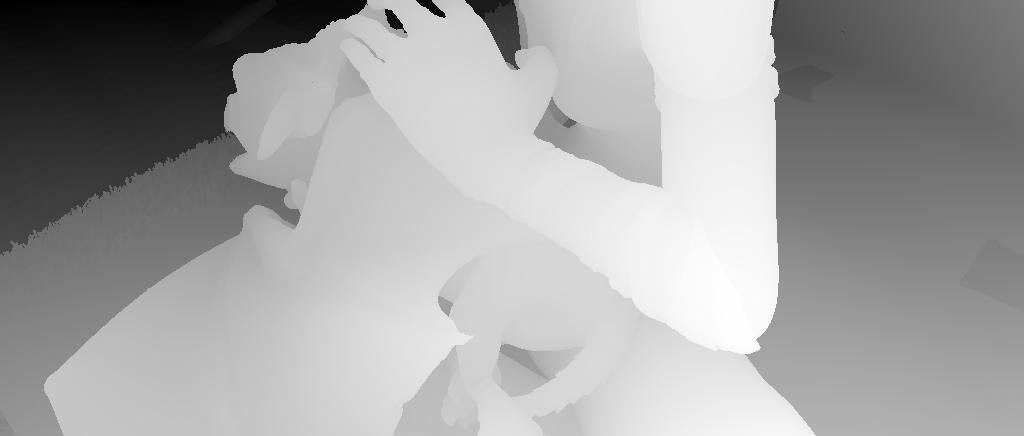}}\hfill
		\subfigure[]
		{\includegraphics[width=0.247\linewidth]{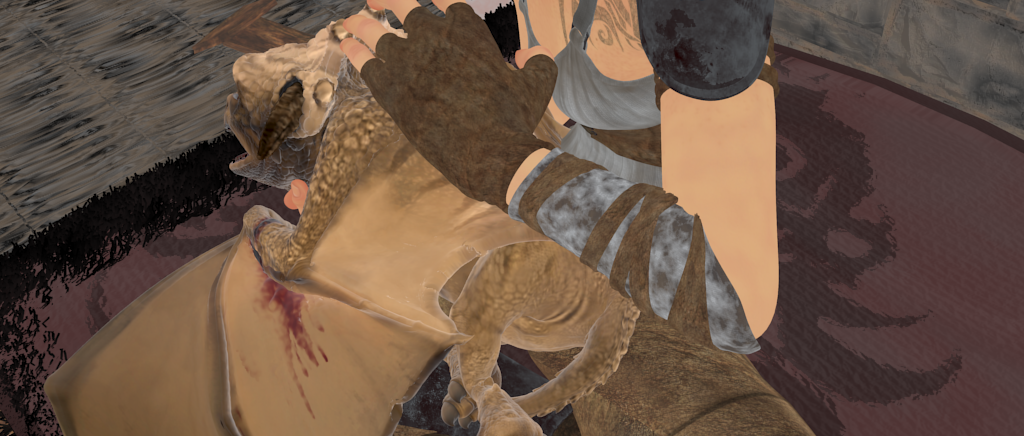}}\hfill
		\subfigure[]
		{\includegraphics[width=0.247\linewidth]{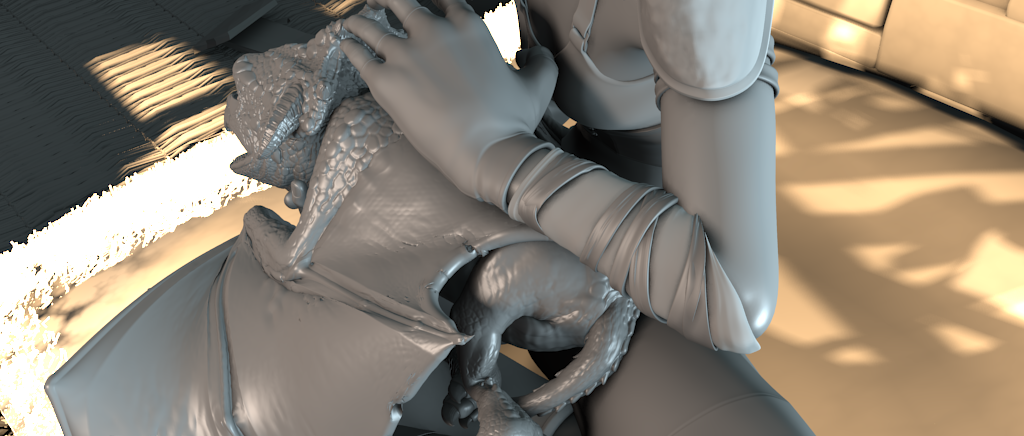}}\hfill
		\vspace{-23pt}
		\subfigure[(a)]
		{\includegraphics[width=0.247\linewidth]{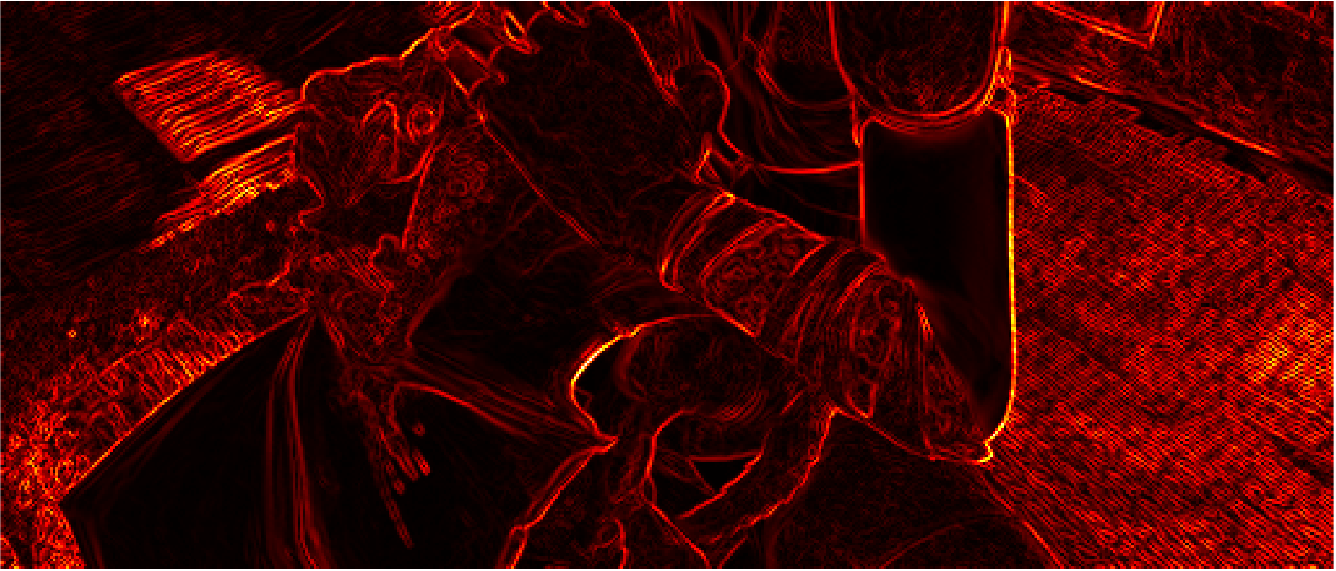}}\hfill
		\subfigure[(b)]
		{\includegraphics[width=0.247\linewidth]{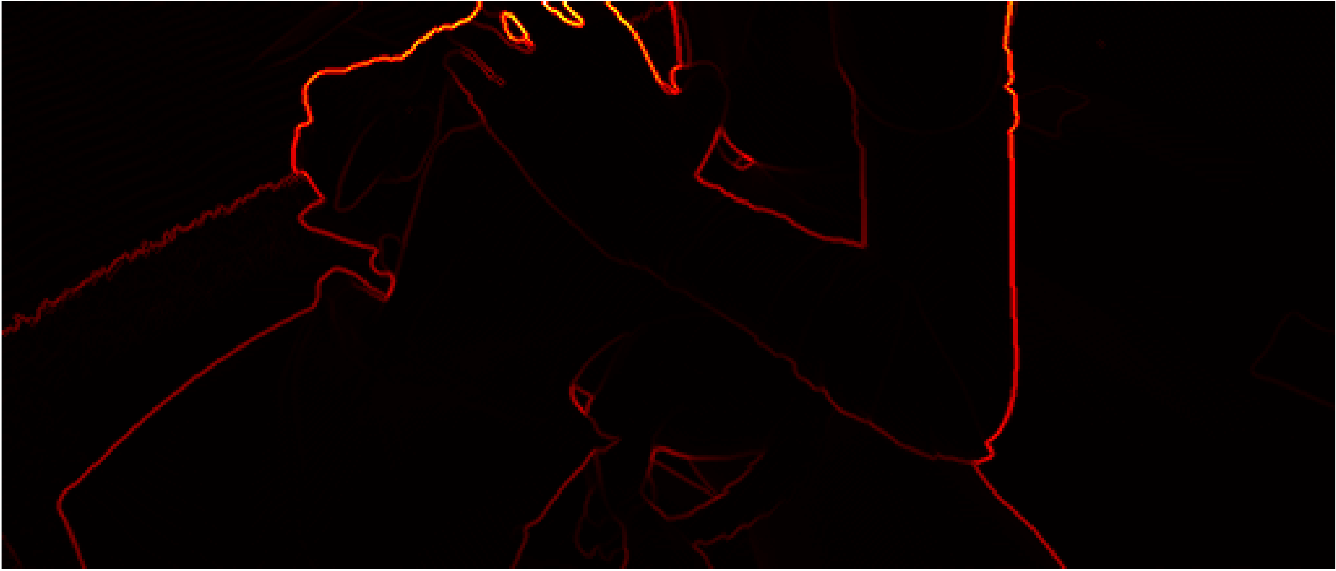}}\hfill
		\subfigure[(c)]
		{\includegraphics[width=0.247\linewidth]{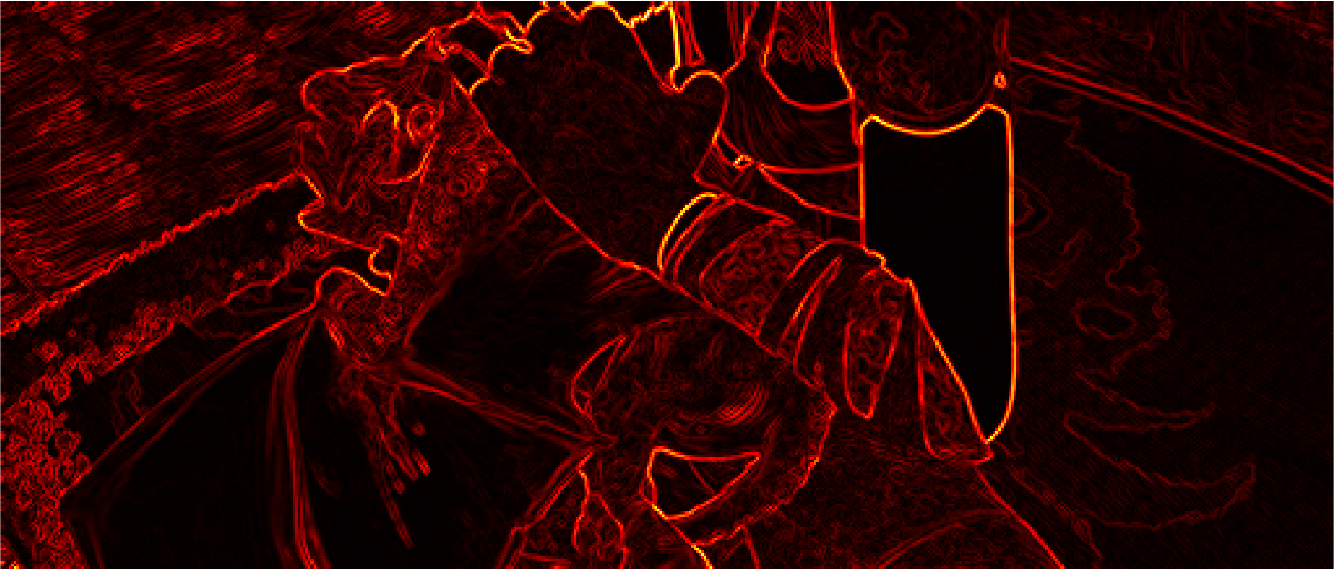}}\hfill
		\subfigure[(d)]
		{\includegraphics[width=0.247\linewidth]{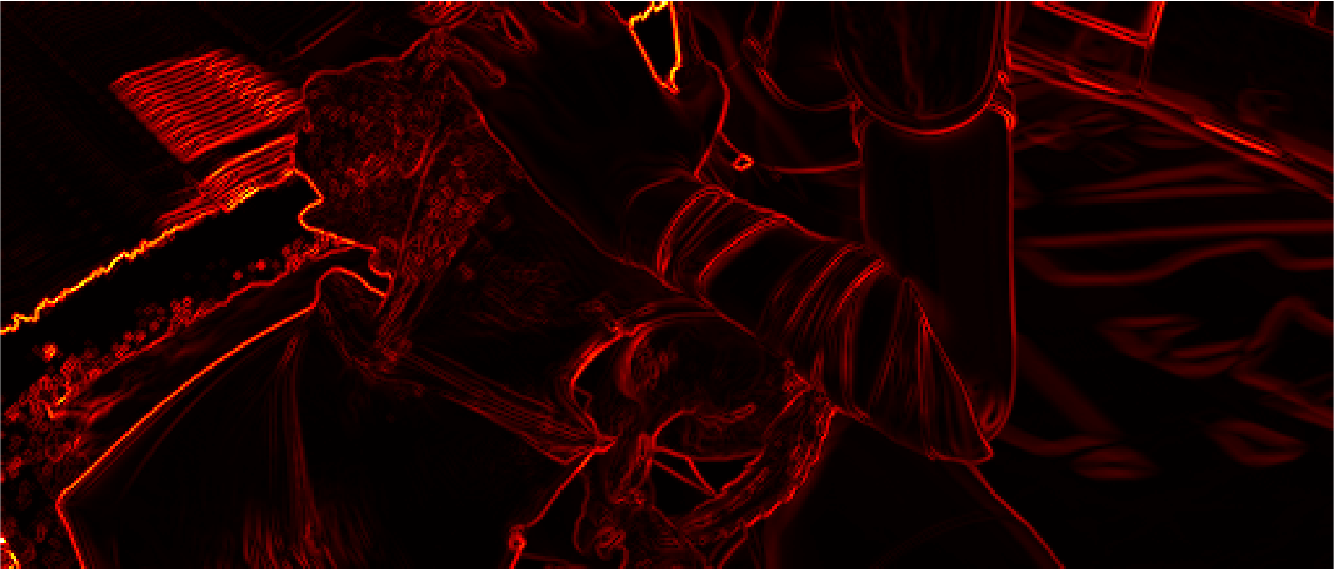}}\hfill
		\vspace{-12pt}
		\caption{For an example from the MPI-SINTEL dataset \cite{Butler12}, its (a) color image $I$, (b) depth $D$, (c) albedo $A$, (d) shading $S$, and their corresponding gradient fields $\triangledown I$, $\triangledown D$, $\triangledown A$, and $\triangledown S$ shown below. Compared to quantities in the value domain, correlations are stronger among gradient fields, such that estimates of one may help in learning others.
			Furthermore, the gradient consistency between $\triangledown I$, $\triangledown D$, $\triangledown A$, and $\triangledown S$ can be used to estimate the confidence of each gradient.}\label{img:10}\vspace{-10pt}
	\end{figure}

	We specifically learn the joint prediction model in the gradient domain, where depth and intrinsic images generally exhibit stronger correlation
	than in the value domain, as exemplified in \figref{img:10}. This greater correlation and reduced discrepancy among $\triangledown D$, $\triangledown A$, and $\triangledown S$ facilitate joint learning of the two tasks by allowing them to better leverage information from each other\footnote{$\triangledown$ is a differential operator defined in the $\mathbf{x}$- and $\mathbf{y}$-direction such that $\triangledown = [\triangledown_{\mathbf{x}},\triangledown_{\mathbf{y}}]$.}. We therefore formulate our model to predict the depth, albedo, and shading gradient fields from the color image. Our method additionally learns the confidence of predicted gradients based on their consistency among one another in the training set.
	
	We formulate this joint prediction using convolutional neural networks (CNNs) in a joint conditional random field (CRF). Our system architecture is structured as three cooperating networks, namely a depth prediction network, an intrinsic prediction network, and a gradient scale network. The depth prediction network is modeled by two feed-forward processes $\mathcal{F} (I^i;\mathbf{w}_\mathcal{F}^{D})$ and $\mathcal{F} (I^i;\mathbf{w}_\mathcal{F}^{\triangledown D})$, where $\mathbf{w}_\mathcal{F}^{D}$ and $\mathbf{w}_\mathcal{F}^{\triangledown D}$ represent the network parameters for depth and depth gradients. The intrinsic prediction network is similarly modeled by feed-forward processes $\mathcal{F} (I^i;\mathbf{w}_\mathcal{F}^{\triangledown A})$ and $\mathcal{F} (I^i;\mathbf{w}_\mathcal{F}^{\triangledown S})$, where $\mathbf{w}_\mathcal{F}^{\triangledown A}$ and $\mathbf{w}_\mathcal{F}^{\triangledown S}$ represent the network parameters for albedo gradients and shading gradients. The gradient scale network learns the confidence of depth, albedo and shading gradients using a feed-forward process for each, denoted by $\mathcal{G} (\triangledown I^i,\triangledown A^i,\triangledown S^i;\mathbf{w}_\mathcal{G}^{\triangledown D})$,
	$\mathcal{G} (\triangledown I^i,\triangledown D^i,\triangledown S^i;\mathbf{w}_\mathcal{G}^{\triangledown A})$, and
	$\mathcal{G} (\triangledown I^i,\triangledown D^i,\triangledown A^i;\mathbf{w}_\mathcal{G}^{\triangledown S})$,
	where $\mathbf{w}_\mathcal{G}^{\triangledown D}$, $\mathbf{w}_\mathcal{G}^{\triangledown A}$, and $\mathbf{w}_\mathcal{G}^{\triangledown S}$ are their respective network parameters. The three networks in our system are jointly learned in a manner where
	each can leverage information from the other networks. \vspace{-5pt}
	
	\subsection{Joint Conditional Random Field}\label{sec:32}
	The networks in our model are jointly learned by minimizing the energy function of a joint CRF. The joint CRF is formulated so that each task can leverage information from the other complementary task, leading to improved prediction in comparison to separate estimation models. Our energy function $\mathbf{E}(D,A,S|I)$ is defined as unary potentials $\mathbf{E}_u$ and pairwise potentials $\mathbf{E}_s$ for each task:
	\begin{equation}\label{equ:energy}
	\begin{split}
	\mathbf{E}(D,A,S|&I) = \mathbf{E}_u(D|I) + \mathbf{E}_u(A,S|I)\\
	&+ \lambda_D \mathbf{E}_{s}(D|I,A,S) + \lambda_A \mathbf{E}_{s}(A|I,D,S) + \lambda_S \mathbf{E}_{s}(S|I,D,A),
	\end{split}
	\end{equation}
	where $\lambda_D$, $\lambda_A$, and $\lambda_S$ are weights for each pairwise potential. In the training procedure, this energy function is minimized over all the training images, \emph{i.e.,} by minimizing $\sum\nolimits_{i} \mathbf{E}(D^i,A^i,S^i|I^i)$. For testing, given a query image $I$ and the learned network parameters, the final solutions of $D$, $A$, and $S$ are estimated by minimizing the energy function $\mathbf{E}(D,A,S|I)$. \vspace{-8pt}
	
	\subsubsection{Unary Potentials} The unary potentials consist of two energy functions, $\mathbf{E}_u(D|I)$ and $\mathbf{E}_u(A,S|I)$. The depth unary function $\mathbf{E}_u(D|I)$ is formulated as
	\begin{equation}\label{equ:unary_depth}
	\mathbf{E}_u (D|I) = \sum\nolimits_{p} { \left( D_{p} - \mathcal{F} (I_{\mathcal{P}};\mathbf{w}_\mathcal{F}^{D}) \right)^2 },
	\end{equation}
	which represents the squared differences between depths $D_p$ and a predicted depths from $\mathcal{F} (I_{\mathcal{P}};\mathbf{w}_\mathcal{F}^{D})$, where $\mathcal{P}$ is the local neighborhood\footnote{It is defined as the receptive field through the CNNs for pixel $p$ \cite{He15}.} for pixel $p$. It can be considered as a Dirichlet boundary condition for depth pairwise potentials, which will be described shortly.
	
	The unary function $\mathbf{E}_u(A,S|I)$ for intrinsic images is used in minimizing the reconstruction errors of color image $I$ from albedo $A$ and shading $S$:
	\begin{equation}\label{equ:unary_intrinsic}
	\mathbf{E}_u(A,S|I) = \sum\nolimits_{p} { \left( L_{p} (I_{p}- A_{p} - S_{p}) \right)^2 },
	\end{equation}
	where $L_{p} = \mathrm{lum}(I_{p})+\varepsilon$,
	and $\mathrm{lum}(I)$ denotes the luminance of $I$ with $\varepsilon=0.001$.
	It has been noted that processing of luminance balances out the influence of the
	unary potential across the image \cite{Chen13,Kong14}, and that treating the
	image formation equation (\emph{i.e.,} $I_{p} = A_{p} + S_{p}$) as a
	soft constraint can bring greater stability in optimization \cite{Bonneel14},
	especially for dark pixels whose chromaticity can be greatly distorted by sensor noise. \vspace{-8pt}
	
	\subsubsection{Pairwise Potentials} The pairwise potentials,
	which include $\mathbf{E}_{s}(D|I,A,S)$, $\mathbf{E}_{s}(A|I,D,S)$, and
	$\mathbf{E}_{s}(S|I,D,A)$, represent differences between
	gradients and estimated gradients in the depth, albedo, and shading images.
	The pairwise potential $\mathbf{E}_{s}(D|I,A,S)$ for depth gradients
	is defined as
	\begin{equation}\label{equ:Pairwise_depth}
	\mathbf{E}_s(D|I,A,S) = \sum\nolimits_{p}{\| \triangledown D_{p} -
		\mathcal{G} (\triangledown I_{\mathcal{P}},\triangledown A_{\mathcal{P}},\triangledown S_{\mathcal{P}};\mathbf{w}_\mathcal{G}^{\triangledown D}) \circ
		\mathcal{F} (I_{\mathcal{P}};\mathbf{w}_\mathcal{F}^{\triangledown D}) \|^2},
	\end{equation}
	where $\circ$ denotes the Hadamard product, and the estimated depth gradients of $\mathcal{F} (I_{p};\mathbf{w}_\mathcal{F}^{\triangledown D})$ provide a guidance gradient field for depth, similar to a Poisson equation \cite{Perez03,Xu15}. They are weighted by a confidence factor
	$\mathcal{G} (\triangledown I_{\mathcal{P}},\triangledown A_{\mathcal{P}},\triangledown S_{\mathcal{P}};\mathbf{w}_\mathcal{G}^{\triangledown D})$
	learned in the gradient scale network to reduce the impact of erroneous gradients. This gradient scale is similar to the derivative-level confidence employed in \cite{Shen15} for image restoration, except that our gradient scale is learned non-locally with CNNs and different types of guidance images, as later described in \secref{sec:34}. The pairwise potentials for albedo gradients $\mathbf{E}_s(A|I,D,S)$ and shading gradients $\mathbf{E}_s(S|I,D,A)$ are defined in the same manner. Since the gradient scales are jointly estimated with each other task, these pairwise potentials are computed within an iterative solver, which will be described in \secref{sec:41}. \vspace{-5pt}
	\begin{figure}[!t]
		\centering
		\renewcommand{\thesubfigure}{}
		\subfigure[]
		{\includegraphics[width=1\linewidth]{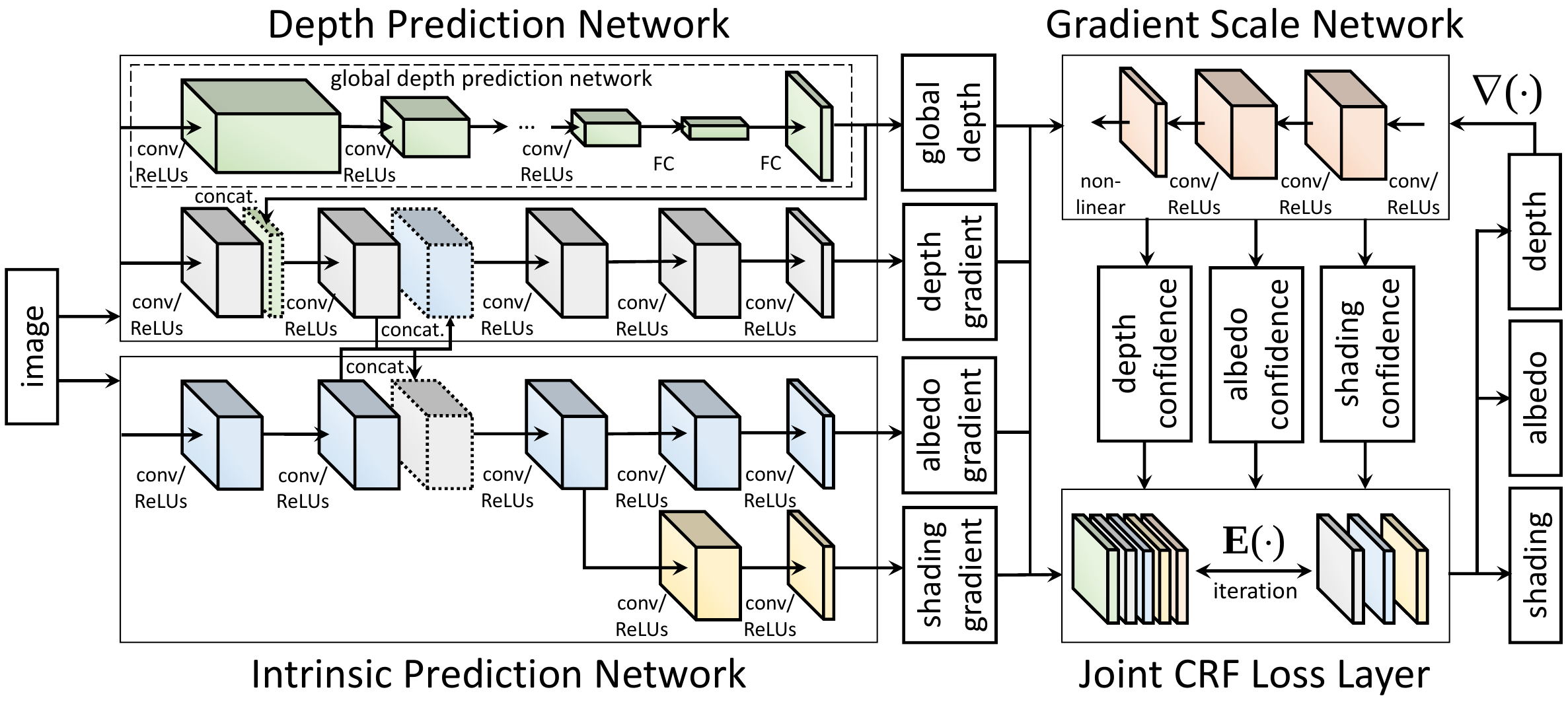}}
		\vspace{-25pt}
		\caption{Network architecture of the JCNF model. It consists of a depth prediction network, an intrinsic prediction network, and a gradient scale network. These networks are learned by minimizing a joint CRF loss function.}\label{img:1}\vspace{-10pt}
	\end{figure}
	
	\subsection{Joint Depth and Intrinsic Prediction Network}\label{sec:33}
	Our joint depth and intrinsic prediction network utilizes the aforementioned energy function to predict $D$, $\triangledown D$, $\triangledown A$, and $\triangledown S$ from a single image $I$. The joint network consists of a depth prediction network for $D$ and $\triangledown D$, and an intrinsic prediction network for $\triangledown A$ and $\triangledown S$. In contrast to previous methods for single-image depth prediction \cite{Eigen14,Eigen15,Narihira15}, our system jointly estimates the gradient fields $\triangledown D$, $\triangledown A$, and $\triangledown S$, which are used to reduce ambiguity in the solution and obtain more edge-preserved results. To allow the different estimation tasks to leverage information from one another, we design the depth and intrinsic networks to share concatenated convolutional activations, and share convolutional layers between albedo and shading networks, as illustrated in \figref{img:1}. \vspace{-8pt}
	
	\subsubsection{Depth Prediction Network}
	The depth prediction network consists of a global depth network and a depth gradient network.
	For the global depth network, we learn its parameters $\mathbf{w}_\mathcal{F}^{D}$ for predicting an overall depth map from the entire image structure.
	Similar to \cite{Eigen14,Eigen15,Narihira15}, it provides coarse,
	spatially-varying depth that may be lacking in fine detail. This coarse
	depth will later be refined using the output of the depth gradient network.
	
	The global depth network consists of five convolutional layers, three pooling layers, six non-linear activation layers, and two fully-connected (FC) layers.
	For the first five layers, the pre-trained parameters from the AlexNet architecture \cite{Alex12} are employed,
	and fine-tuning for the dataset is done.
	Rectified linear units (ReLUs) are used for the non-linear layers, and
	the pooling layers employ max pooling.
	The first FC layer encodes the network responses 
	into fixed-dimensional features,
	and the second FC layer infers a coarse global depth map
	at $1/16$-scale of the original depth map.
	
	The depth gradient network predicts fine-detail depth gradients for each pixel.
	Its parameters $\mathbf{w}_\mathcal{F}^{\triangledown D}$
	are learned using an end-to-end patch-level scheme inspired by \cite{Dong15,Xu15}, where the network input is an image patch and the output is a depth gradient patch.
	For inference of depth gradients at the pixel level,
	the depth gradient network consists of five convolutional networks
	followed by ReLUs, without stride convolutions or pooling layers.
	The first convolutional layer is identical to the first convolutional layer
	in the AlexNet architecture \cite{Alex12}.
	Four additional convolutional layers are also used as shown in \figref{img:1}. The depth gradient patches that are output by this network will be used for depth reconstruction in \secref{sec:42}.
	Note that in the testing procedure,
	the depth gradient network is applied to overlapping patches over the entire image, which are aggregated in the last convolutional layer to yield the full gradient field. \vspace{-8pt}
	\begin{table}[t]
		\centering
		\begin{tabular}{| >{\centering}m{0.09\linewidth}
				| >{\centering}m{0.08\linewidth}  | >{\centering}m{0.08\linewidth}
				| >{\centering}m{0.08\linewidth}  | >{\centering}m{0.08\linewidth}
				| >{\centering}m{0.08\linewidth}  | >{\centering}m{0.08\linewidth}
				| >{\centering}m{0.08\linewidth}  | >{\centering}m{0.08\linewidth}
				| >{\centering}m{0.08\linewidth}  | >{\centering}m{0.08\linewidth} |}
			\hline
			& \multicolumn{7}{ c |}{global depth net.}
			& \multicolumn{3}{ c |}{gradient scale net.} \tabularnewline
			\cline{2-11}
			&conv1 &conv2 &conv3 &conv4 &conv5 &FC1 &FC2 &conv1 &conv2 &conv3 \tabularnewline
			\hline
			kernel &$11\times11$ &$5\times5$ &$3\times3$ &$3\times3$ &$3\times3$ &$1\times1$ &$1\times1$ &$3\times3$ &$3\times3$ &$1\times1$ \tabularnewline
			channel &$96$ &$256$ &$384$ &$384$ &$256$ &$4096$ &-$/16$ &$64$ &$64$ &$2$ or $6$ \tabularnewline
			\hline
			\hline
			& \multicolumn{5}{ c |}{depth gradient net.}
			& \multicolumn{5}{ c |}{intrinsic gradient net.} \tabularnewline
			\cline{2-11}
			&conv1 &conv2 &conv3 &conv4 &conv5 &conv1 &conv2 &conv3 &conv4 &conv5 \tabularnewline
			\hline
			kernel &$11 \times 11$ &$3\times3$ &$3\times3$ &$3\times3$ &$3\times3$ &$11 \times 11$ &$3\times3$ &$3\times3$ &$3\times3$ &$3\times3$ \tabularnewline
			channel &$96+1$ &$64+64$ &$64$ &$64$ &$2$ &$96$ &$64+64$ &$64$ &$64$ &$6$ \tabularnewline
			\hline
		\end{tabular}\vspace{+3pt}
		\caption{Network architecture of the JCNF model.}\label{tab:0}\vspace{-20pt}
	\end{table}
	
	\subsubsection{Intrinsic Prediction Network}
	The intrinsic prediction network has a structure similar to the
	depth gradient prediction network.
	The network parameters
	$\mathbf{w}_\mathcal{F}^{\triangledown A}$ and $\mathbf{w}_\mathcal{F}^{\triangledown S}$
	are learned for predicting the albedo and shading gradients at each pixel.
	To jointly infer the depth and intrinsic image gradients,
	the second convolutional activations for each task are concatenated and passed to their third convolutional layers as shown in \figref{img:1}.
	In the training procedure, the depth and intrinsic networks are iteratively learned, which enables each task to benefit from each other's activations to provide more reliable estimates.
	Furthermore, similar to \cite{Narihira15}, the albedo and shading gradient networks
	share their first three convolutional layers, while the last two are separate.
	Since the albedo and shading images have related properties,
	these shared convolutional layers benefit their estimation. Details on kernel sizes and the number of channels for each layer are provided in \tabref{tab:0} for all the networks. \vspace{-5pt}
	
	\subsection{Gradient Scale Network}\label{sec:34}
	The estimated gradients from the depth and intrinsic prediction networks
	might contain errors due to the ill-posed nature of their problems.
	To help in identifying such errors, our system additionally
	learns the confidence of estimated gradients, specifically, whether a gradient
	exists at a particular location or not.
	The basic idea is to learn from the training data about the
	consistencies that exist among the different types of gradients given their local neighborhood $\mathcal{P}$.
	From this, we can determine the confidence of a gradient ({\it e.g.},
	a depth gradient), based on the other estimated gradients ({\it e.g.},
	the albedo, shading, and image gradients).
	This confidence is modeled as a gradient scale that is similar to
	the scale map used in \cite{Shen15} to model derivative-level confidence
	for image restoration.
	It can be noted that in some depth and intrinsic image decomposition methods \cite{Chen13,Jeon14,Fayao15},
	the solutions are filtered with fixed parameters using the color image as guidance.
	Our system instead learns a network for defining the parameters,
	using not only a color image but also depth and intrinsic images as guidance.
	
	The gradient scale network consists of three convolutional layers and one non-linear activation layer.
	For the case of depth gradients, the output of the gradient scale network $\mathcal{G} (\triangledown I_{\mathcal{P}},\triangledown A_{\mathcal{P}},\triangledown S_{\mathcal{P}};\mathbf{w}_\mathcal{G}^{\triangledown D})$ is estimated as
	the convolution between $\mathbf{w}_\mathcal{G}^{\triangledown D}$ and $(|\triangledown I_{\mathcal{P}}|^2,|\triangledown A_{\mathcal{P}}|^2,|\triangledown S_{\mathcal{P}}|^2)$, followed by a non-linear activation \emph{i.e.,}
	$f(\cdot)=(1-\mathrm{\mathbf{exp}}(1-\cdot))/(1+\mathrm{\mathbf{exp}}(1-\cdot))$, which is defined within $[-1,1]$. Here, $|\cdot|^2$ for a vector of gradients denotes a vector of the gradient magnitudes. Thus, in the gradient scale network, the network parameters are convolved with the gradient magnitudes.
	With the learned parameters $\mathbf{w}_\mathcal{G}^{\triangledown D}$,
	the confidence of $\triangledown D_{p}$ is estimated from $\triangledown I_{\mathcal{P}}$, $\triangledown A_{\mathcal{P}}$, $\triangledown S_{\mathcal{P}}$. This can alternatively be viewed as a guidance filtering weight for $D$ with guidance images $I$, $A$, and $S$. $\mathcal{G} (\triangledown I_\mathcal{P},\triangledown D_\mathcal{P},\triangledown S_\mathcal{P};\mathbf{w}_\mathcal{G}^{\triangledown A})$ and
	$\mathcal{G} (\triangledown I_\mathcal{P},\triangledown D_\mathcal{P},\triangledown A_\mathcal{P};\mathbf{w}_\mathcal{G}^{\triangledown S})$ are also similarly defined.
	
	Some properties of gradient scales are as follows.
	A gradient scale can be either positive or negative.
	A large positive value indicates high confidence in the presence of a gradient.
	A large negative value also indicates high confidence,
	but for the reversed gradient direction.
	In addition, when a gradient field contains extra erroneous regions,
	gradient scales of value $0$ can help to disregard them. \vspace{-5pt}
	
	\section{Unified Depth and Intrinsic Image Prediction}\label{sec:4}
	\vspace{-5pt}
	\subsection{Training}\label{sec:41}
	The energy function $\mathbf{E}(D,A,S|I)$ from \equref{equ:energy} is used to simultaneously learn the depth and intrinsic network parameters ($\mathbf{w}_\mathcal{F}^{D}$, $\mathbf{w}_\mathcal{F}^{\triangledown D}$,
	$\mathbf{w}_\mathcal{F}^{\triangledown A}$, $\mathbf{w}_\mathcal{F}^{\triangledown S}$) and the gradient scale network parameters ($\mathbf{w}_\mathcal{G}^{\triangledown D}$, $\mathbf{w}_\mathcal{G}^{\triangledown A}$, $\mathbf{w}_\mathcal{G}^{\triangledown S}$).
	Although the overall form of the energy is non-quadratic, it has a quadratic form with respect to each of its terms. The energy function can thus be minimized by alternating among its terms. \vspace{-8pt}
	
	\subsubsection{Loss Functions}
	For the global depth unary potential of \equref{equ:unary_depth}, the global depth network parameters $\mathbf{w}_\mathcal{F}^{D}$ can be solved by minimizing the following loss function
	\begin{equation}\label{equ:Pairwise_globdepth_loss}
	\mathcal{L}(\mathbf{w}_\mathcal{F}^{D}) = \sum\nolimits_{\{i,p\}} { \left( D^i_{p} - \mathcal{F} (I^i_{\mathcal{P}};\mathbf{w}_\mathcal{F}^{D}) \right)^2 }.
	\end{equation}
	
	We note that the intrinsic image unary term does not contain network parameters to be learned, so it is used only in the testing procedure.
	
	The pairwise potentials each incorporate two networks, namely the gradient prediction network and gradient scale network, so they are iteratively trained. The loss function for the depth gradient pairwise potential of (\ref{equ:Pairwise_depth}) is defined as
	\begin{equation}\label{equ:Pairwise_depth_loss}
	\mathcal{L}(\mathbf{w}_\mathcal{G}^{\triangledown D},\mathbf{w}_\mathcal{F}^{\triangledown D}) = \sum\nolimits_{\{i,p\}} {\| \triangledown D^i_{p} -
		\mathcal{G}(\triangledown I^i_{\mathcal{P}},\triangledown A^i_{\mathcal{P}},\triangledown S^i_{\mathcal{P}};\mathbf{w}_\mathcal{G}^{\triangledown D}) \circ \mathcal{F} (I^i_{\mathcal{P}};\mathbf{w}_\mathcal{F}^{\triangledown D}) \|^2}.
	\end{equation}
	The loss functions for the pairwise potentials of the albedo gradients $\mathcal{L}(\mathbf{w}_\mathcal{G}^{\triangledown A},\mathbf{w}_\mathcal{F}^{\triangledown A})$ and shading gradients $\mathcal{L}(\mathbf{w}_\mathcal{G}^{\triangledown S},\mathbf{w}_\mathcal{F}^{\triangledown S})$ are similarly defined.
	
	These loss functions are minimized using stochastic gradient descent with the standard back-propagation \cite{MatConv}. 
	First, $\mathbf{w}_\mathcal{F}^{D}$ is estimated through $\partial \mathcal{L}(\mathbf{w}_\mathcal{F}^{D}) / \partial \mathbf{w}_\mathcal{F}^{D}$.
	Then $\mathbf{w}_\mathcal{G}^{\triangledown D}$ and $\mathbf{w}_\mathcal{F}^{\triangledown D}$ are iteratively estimated through $\partial \mathcal{L}(\mathbf{w}_\mathcal{G}^{\triangledown D},\mathbf{w}_\mathcal{F}^{\triangledown D}) / \partial \mathbf{w}_\mathcal{G}^{\triangledown D}$ and $\partial \mathcal{L}(\mathbf{w}_\mathcal{G}^{\triangledown D},\mathbf{w}_\mathcal{F}^{\triangledown D}) / \partial \mathbf{w}_\mathcal{F}^{\triangledown D}$.
	In each iteration, the loss functions are differently defined according to the other network outputs, where the network parameters are initialized with the values obtained from the previous iteration. In this way, the networks are trained jointly and account for the improving outputs of the other networks.
	\vspace{-5pt}
	\begin{table}[!t]
		\begin{center}
			\begin{tabularx}{\linewidth}{p{6mm} p{1mm}| p{3mm}| p{160mm}}
				\hlinewd{0.8pt}
				~&\multicolumn{3}{ p{165mm} }{{\bf Algorithm 1}: Unified Depth and Intrinsic Image Prediction}\\
				\hlinewd{0.8pt}
				~&\multicolumn{3}{ p{165mm} }{{\bf Input}: training image set $\mathcal{C} = \{ \left({I^i},{D^i},{A^i},{S^i}\right)\}$, query color image $I^*$}\\
				~&\multicolumn{3}{ p{165mm} }{{\bf Output}: depth $D$, albedo $A$, shading $S$.}\\
				~&\multicolumn{3}{ p{165mm} }{{\bf $/*$ Training Procedure $*/$}}\\
				$\mathbf{1:}$&\multicolumn{3}{ p{165mm} }{For training set $\mathcal{C}$,
					learn parameters $\mathbf{w}_\mathcal{F}^{D}$ using backward-propagation.}\\
				$\mathbf{2:}$&\multicolumn{3}{ p{165mm} }{Initialize parameters of $\mathbf{w}_\mathcal{G}^{\triangledown D}$,
					$\mathbf{w}_\mathcal{G}^{\triangledown A}$,
					and $\mathbf{w}_\mathcal{G}^{\triangledown S}$ to provide constant values.}\\
				~&\multicolumn{3}{ p{165mm} }{{\bf while} not converged {\bf do }}\\
				$\mathbf{3:}$&~&\multicolumn{2}{ p{165mm} }{$\;\;$For $\mathcal{C}$, update parameters $\mathbf{w}_\mathcal{F}^{\triangledown D}$,
				$\mathbf{w}_\mathcal{F}^{\triangledown A}$, $\mathbf{w}_\mathcal{F}^{\triangledown S}$ with fixed $\mathbf{w}_\mathcal{G}^{\triangledown D}$, $\mathbf{w}_\mathcal{G}^{\triangledown A}$, $\mathbf{w}_\mathcal{G}^{\triangledown S}$.}\\
				$\mathbf{4:}$&~&\multicolumn{2}{ p{165mm} }{$\;\;$For $\mathcal{C}$, update parameters $\mathbf{w}_\mathcal{G}^{\triangledown D}$, $\mathbf{w}_\mathcal{G}^{\triangledown A}$, $\mathbf{w}_\mathcal{G}^{\triangledown S}$ with fixed $\mathbf{w}_\mathcal{F}^{\triangledown D}$, $\mathbf{w}_\mathcal{F}^{\triangledown A}$, $\mathbf{w}_\mathcal{F}^{\triangledown S}$.}\\
				~&\multicolumn{3}{ p{165mm} }{{\bf end while }}\\
				~&\multicolumn{3}{ p{165mm} }{{\bf $/*$ Testing Procedure $*/$}}\\
				~&\multicolumn{3}{ p{165mm} }{{\bf for } $l=1:\mathcal{N}_L$ {\bf do }}\\
				$\mathbf{5:}$&~&\multicolumn{2}{ p{165mm} }{$\;\;$Estimate $D^{l,*}$, $\triangledown D^{l,*}$ using forward-propagation $\mathcal{F} (I^l;\mathbf{w}_\mathcal{F}^{D})$, $\mathcal{F} (I^l;\mathbf{w}_\mathcal{F}^{\triangledown D})$.}\\
				$\mathbf{6:}$&~&\multicolumn{2}{ p{165mm} }{$\;\;$Estimate $\triangledown A^{l,*}$, $\triangledown S^{l,*}$ using forward-propagation $\mathcal{F} (I^l;\mathbf{w}_\mathcal{F}^{\triangledown A})$, $\mathcal{F} (I^l;\mathbf{w}_\mathcal{F}^{\triangledown S})$.}\\
				~&~&\multicolumn{2}{ p{165mm} }{$\;\;${\bf while} not converged {\bf do }}\\
				$\mathbf{7:}$&~&~&$\;\;$Estimate $C ({\triangledown D^{l,*}})$, $C ({\triangledown A^{l,*}})$, and $C ({\triangledown S^{l,*}})$ using forward-propagation.\\
				$\mathbf{8:}$&~&~&$\;\;$Estimate $D^{l}$, $A^{l}$, $S^{l}$ by optimizing $\mathbf{E}(D^{l}|I^{l},I^{l-1})$ and $\mathbf{E}(A^{l},S^{l}|I^{l},I^{l-1})$.\\
				$\mathbf{9:}$&~&~&$\;\;$Compute $D^{l,*}$, $\triangledown D^{l,*}$, $\triangledown A^{l,*}$, $\triangledown S^{l,*}$ from $D^{l}$, $A^{l}$, $S^{l}$. \\
				~&~&\multicolumn{2}{ p{165mm} }{$\;\;${\bf end while }}\\
				$\mathbf{10:}$&~&\multicolumn{2}{ p{165mm} }{$\;\;$Interpolate $D^{l}$, $A^{l}$, $S^{l}$ into the size of $I^{l+1}$ using a bilinear interpolation.}\\
				~&\multicolumn{3}{ p{165mm} }{{\bf end for }}\\
				$\mathbf{11:}$&\multicolumn{3}{ p{165mm} }{Estimate depth $D$, albedo $A$, shading $S$ as $D^{\mathcal{N}_L}$, $A^{\mathcal{N}_L}$, $S^{\mathcal{N}_L}$.}\\
				\hlinewd{0.8pt}
			\end{tabularx}
		\end{center}\label{alg:1}\vspace{-20pt}
	\end{table}
	
	\subsection{Testing}\label{sec:42}
	\subsubsection{Iterative Joint Prediction}
	In the testing procedure, the outputs $D$, ${\triangledown D}$, ${\triangledown A}$ and ${\triangledown S}$ for a given input image $I$ are predicted by minimizing the energy function $\mathbf{E}(D,A,S|I)$ from \equref{equ:energy} with constraints from the estimates computed using the learned network parameters and forward-propagation.
	Similar to the training procedure, we minimize $\mathbf{E}(D,A,S|I)$ with an iterative scheme due to its non-quadratic form, where $\mathbf{E}(D|I)$ and $\mathbf{E}(A,S|I)$ are minimized in alternation.
	
	For the depth prediction, $\mathbf{E}(D|I)$ is defined as a data term for global depth
	and a pairwise term for depth gradients:
	\begin{equation}\label{equ:energy_depth}
	\mathbf{E}(D|I) = \sum\nolimits_{p} { \left( D_{p} - D^*_{p} \right)^2 }
	+ \lambda_D \sum\nolimits_{p}{\| \triangledown D_{p} - C ({\triangledown D^*_p}) \circ \triangledown D^*_{p} \|^2},
	\end{equation}
	where $*$ denotes network outputs, and $C ({\triangledown D^*_p})$ is the gradient scale of ${\triangledown D^*_p}$ derived from $\mathcal{G} (\triangledown I^{*}_{\mathcal{P}},\triangledown A^{*}_{\mathcal{P}},\triangledown S^{*}_{\mathcal{P}};\mathbf{w}_\mathcal{G}^{\triangledown D})$.
	We note that since $C ({\triangledown D^*_p})$ is computed with $\triangledown I^{*}_{\mathcal{P}}$, $\triangledown A^{*}_{\mathcal{P}}$, and $\triangledown S^{*}_{\mathcal{P}}$, all of the predictions need to be iteratively estimated.
	
	For the intrinsic prediction, $\mathbf{E}(A,S|I)$ is also defined as data and pairwise terms, with the image formation equation and the albedo and shading gradients:
	\begin{equation}\label{equ:energy_intrinsic}
	\begin{split}
	\mathbf{E}(&A,S|I) = \sum\nolimits_{p} { \left( L_{p} (I_{p}- A_{p} - S_{p}) \right)^2 } \\
	&+ \sum\nolimits_{p} {\lambda_A \| \triangledown A_{p} -
		C ({\triangledown A^*_p}) \circ \triangledown A^*_{p} \|^2 +
		\lambda_S \| \triangledown S_{p} - C ({\triangledown S^*_p}) \circ \triangledown S^*_{p} \|^2},
	\end{split}
	\end{equation}
	where $C ({\triangledown A^*_p})$ and $C ({\triangledown S^*_p})$ are defined similarly to $C ({\triangledown D^*_p})$. This energy function can be optimized with an existing linear solver \cite{Chen13}.
	These two energy functions $\mathbf{E}(D|I)$ and $\mathbf{E}(A,S|I)$ are
	iteratively minimized while providing information in the form of depth, albedo, and shading gradients to each other. \vspace{-8pt}
	
	\subsubsection{Coarse-to-Fine Joint Prediction}
	In estimating depth and intrinsic images, enforcing a degree of global consistency can lead to performance gains \cite{Chen13,Barron13}. Although our JCNF model is solved by global energy minimization, its global consistency is limited because gradients are defined just between pixel neighbors.
	For greater global consistency, we apply our joint prediction model in a coarse-to-fine manner, where color images $I^l$ are constructed at $\mathcal{N}_L$ image pyramid levels $l = \{1,...,\mathcal{N}_L\}$, and the depth $D^l$ and intrinsic images $A^l$ and $S^l$ are predicted from $I^l$. Coarser scale results are then used as guidance for finer levels.
	
	Specifically, we reformulate $\mathbf{E}(D|I)$ as $\mathbf{E}(D^{l}|I^{l},I^{l-1})$:
	\begin{equation}\label{equ:energy_depth}
	\begin{split}
	\mathbf{E}(D^{l}|I^{l},I^{l-1}) = &\sum\nolimits_{p} { \left( D^{l}_{p} - D^{l,*}_{p} \right)^2 }
	+ \sum\nolimits_{p} { \left( D^{l}_{p} - D^{l-1}_{p} \right)^2 } \\
	&+ \lambda_D \sum\nolimits_{p}{\| \triangledown D^{l}_{p} - C ({\triangledown D^{l,*}_p}) \circ \triangledown D^{l,*}_{p} \|^2}.
	\end{split}
	\end{equation}
	
	Similarly, $\mathbf{E}(A,S|I)$ is reformulated as $E(A^{l},S^{l}|I^{l},I^{l-1})$:
	\begin{equation}\label{equ:energy_intrinsic}
	\begin{split}
	\mathbf{E}(&A^{l},S^{l}|I^{l},I^{l-1}) = \sum\nolimits_{p} {( L^{l}_{p} (I^{l}_{p}- A^{l}_{p} - S^{l}_{p}) )^2 + (A^{l}_{p} - A^{l-1}_{p})^2 + (S^{l}_{p} - S^{l-1}_{p})^2} \\
	&+ \sum\nolimits_{p} {\lambda_A \| \triangledown A^{l}_{p} -
		C ({\triangledown A^{l,*}_p}) \circ \triangledown A^{l,*}_{p} \|^2 +
		\lambda_S \| \triangledown S^{l}_{p} - C ({\triangledown S^{l,*}_p}) \circ \triangledown S^{l,*}_{p} \|^2},
	\end{split}
	\end{equation}
	where the multi-scale unary functions lead to more reliable solutions and faster convergence.
	The high-level algorithm for the training and testing procedures is provided in the Algorithm 1.
	\vspace{-5pt}
	\begin{figure}[t]
		\centering
		\renewcommand{\thesubfigure}{}
		\subfigure[]
		{\includegraphics[width=0.197\linewidth]{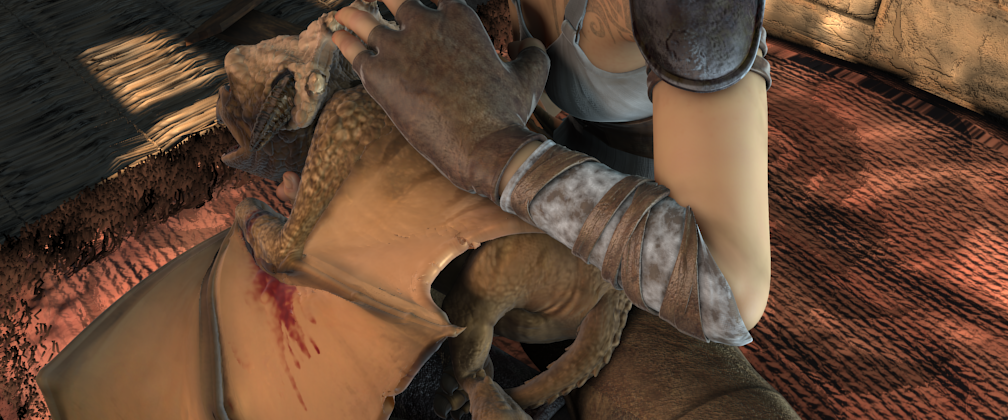}}\hfill
		\subfigure[]
		{\includegraphics[width=0.197\linewidth]{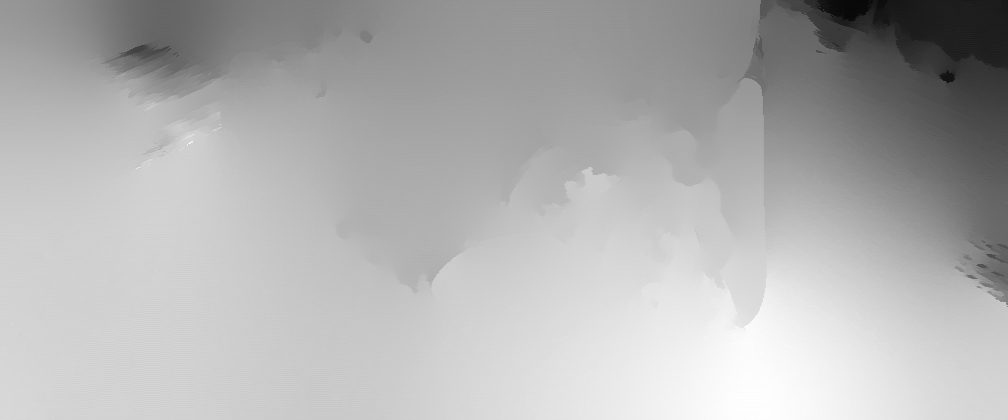}}\hfill
		\subfigure[]
		{\includegraphics[width=0.197\linewidth]{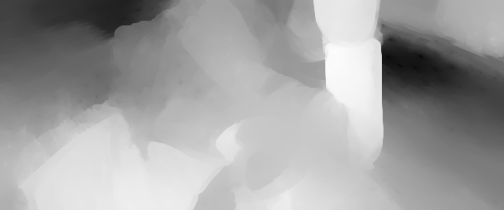}}\hfill
		\subfigure[]
		{\includegraphics[width=0.197\linewidth]{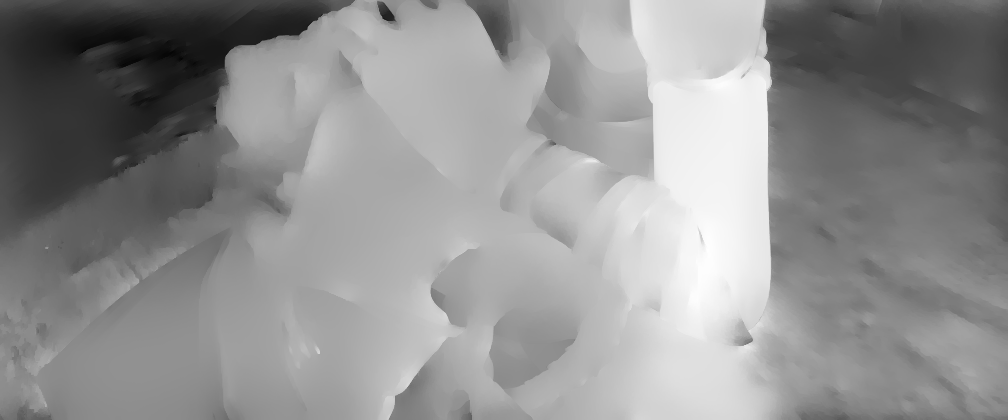}}\hfill
		\subfigure[]
		{\includegraphics[width=0.197\linewidth]{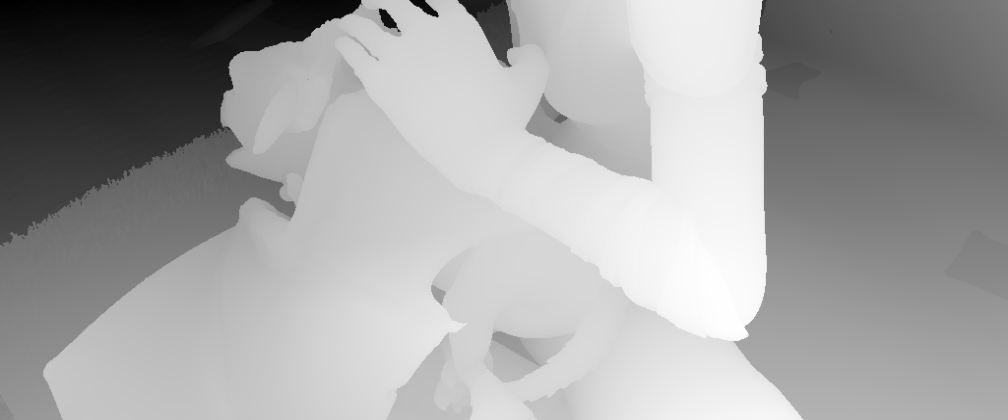}}\hfill
		\vspace{-21pt}
		\subfigure[(a)]
		{\includegraphics[width=0.197\linewidth]{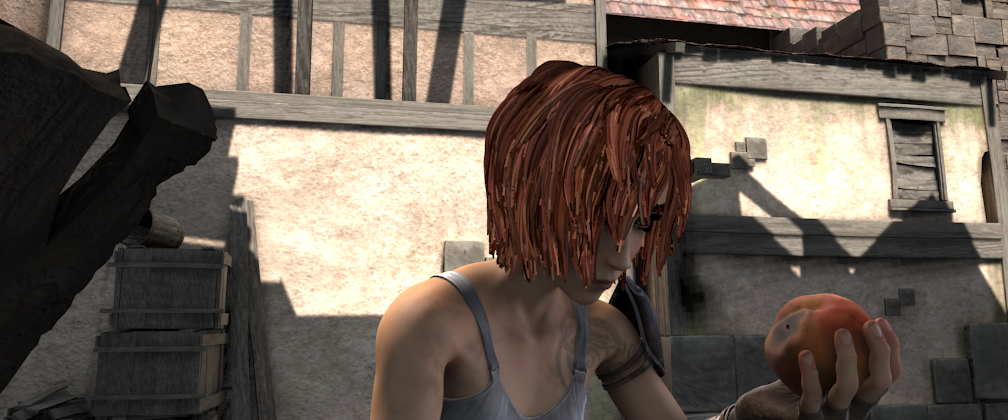}}\hfill
		\subfigure[(b)]
		{\includegraphics[width=0.197\linewidth]{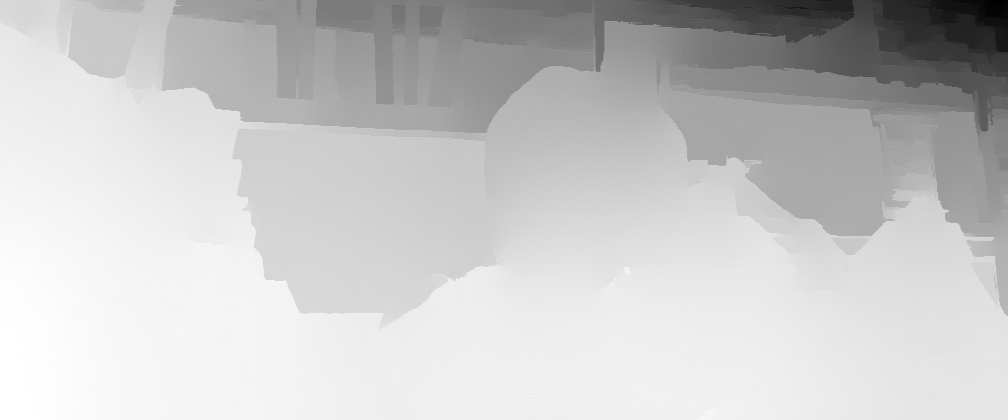}}\hfill
		\subfigure[(c)]
		{\includegraphics[width=0.197\linewidth]{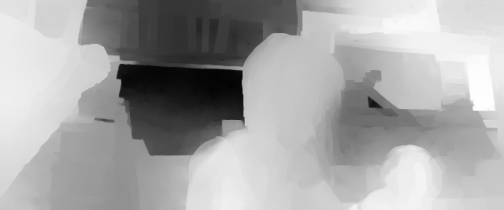}}\hfill
		\subfigure[(d)]
		{\includegraphics[width=0.197\linewidth]{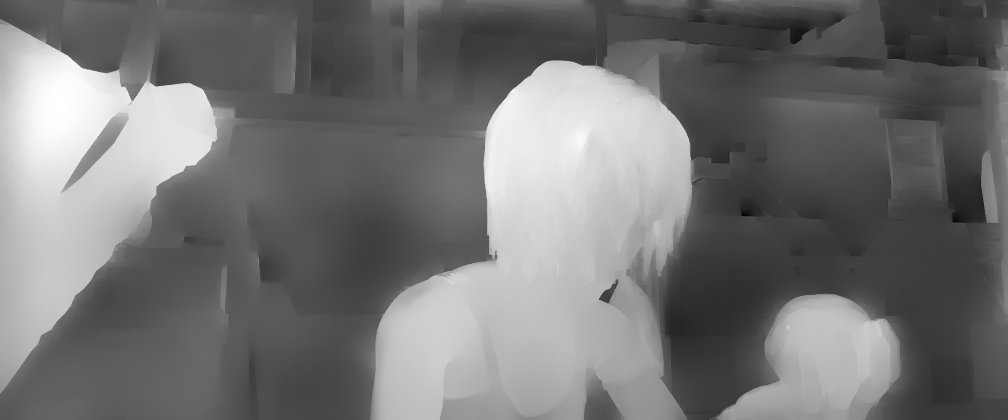}}\hfill
		\subfigure[(e)]
		{\includegraphics[width=0.197\linewidth]{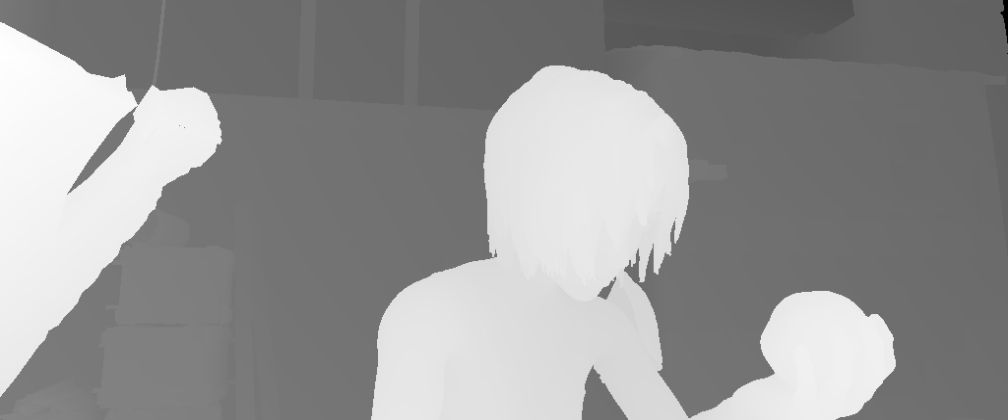}}\hfill
		\vspace{-12pt}
		\caption{Qualitative results on MPI SINTEL \cite{MPIsintel} for depth prediction. (a) color image, (b) DA \cite{Choi15}, (c) DCNF-FCSP{\scriptsize(NYU)} \cite{Fayao15}, (d) JCNF, and (e) ground truth.}\label{img:4}\vspace{-10pt}
	\end{figure}
	\begin{figure}[!t]
		\centering
		\renewcommand{\thesubfigure}{}
		\subfigure[]
		{\includegraphics[width=0.197\linewidth]{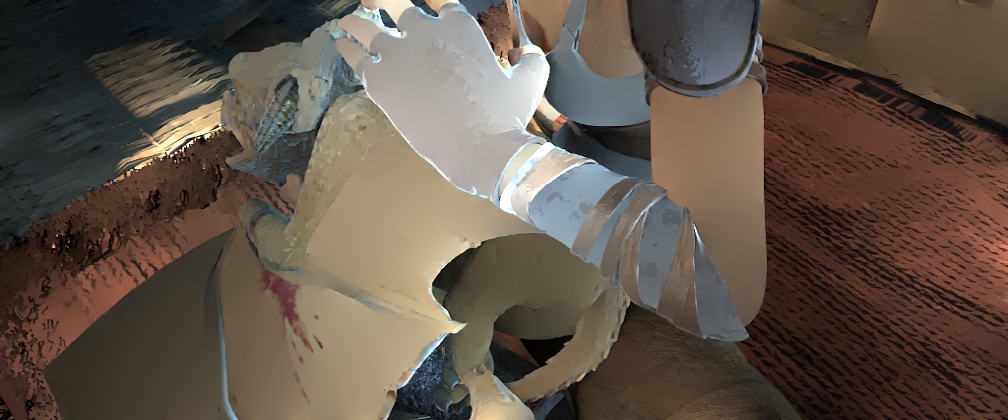}}\hfill
		\subfigure[]
		{\includegraphics[width=0.197\linewidth]{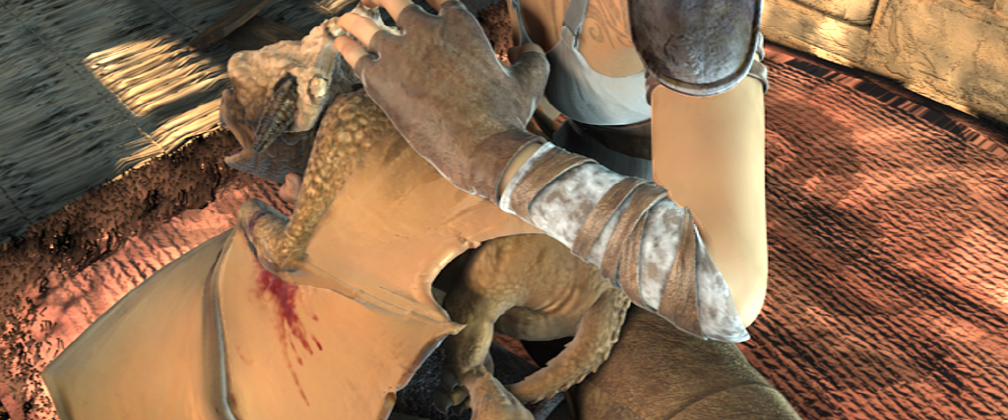}}\hfill
		\subfigure[]
		{\includegraphics[width=0.197\linewidth]{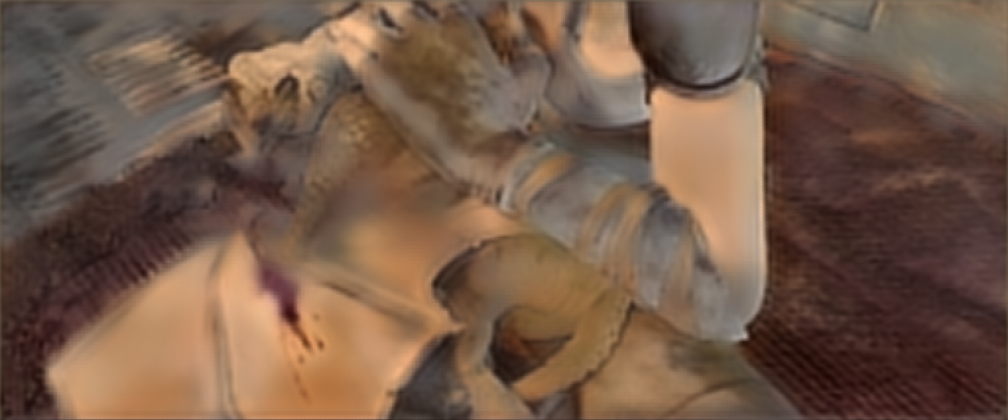}}\hfill
		\subfigure[]
		{\includegraphics[width=0.197\linewidth]{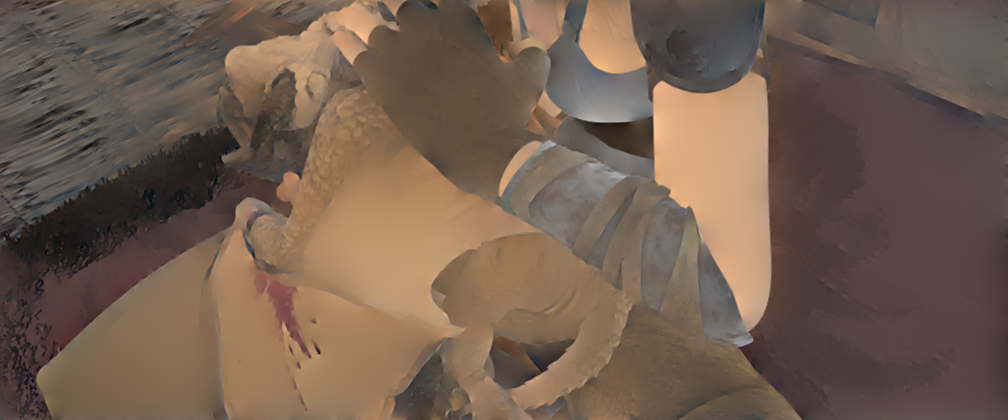}}\hfill
		\subfigure[]
		{\includegraphics[width=0.197\linewidth]{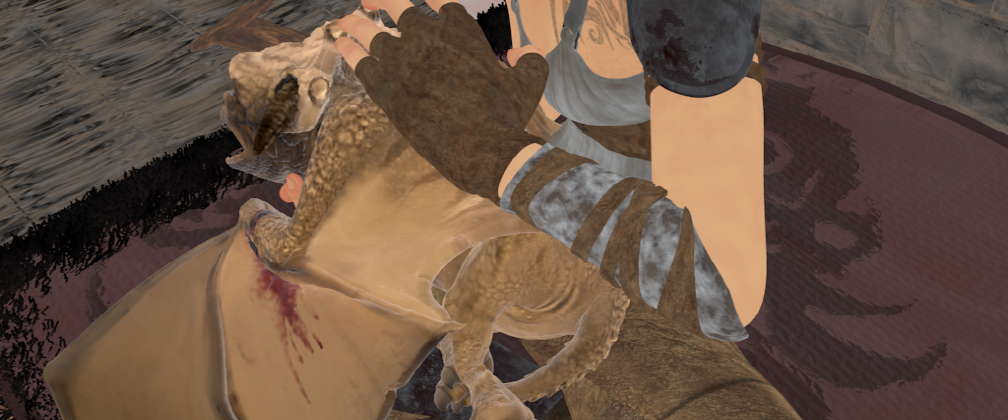}}\hfill
		\vspace{-21pt}
		\subfigure[]
		{\includegraphics[width=0.197\linewidth]{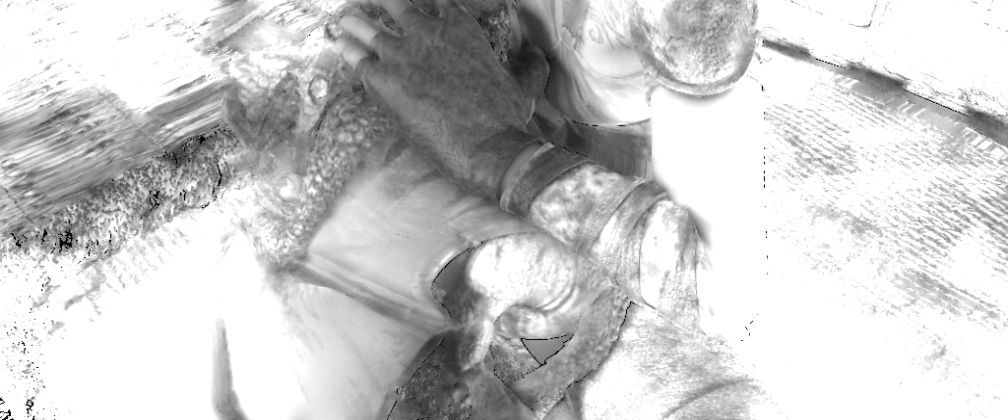}}\hfill
		\subfigure[]
		{\includegraphics[width=0.197\linewidth]{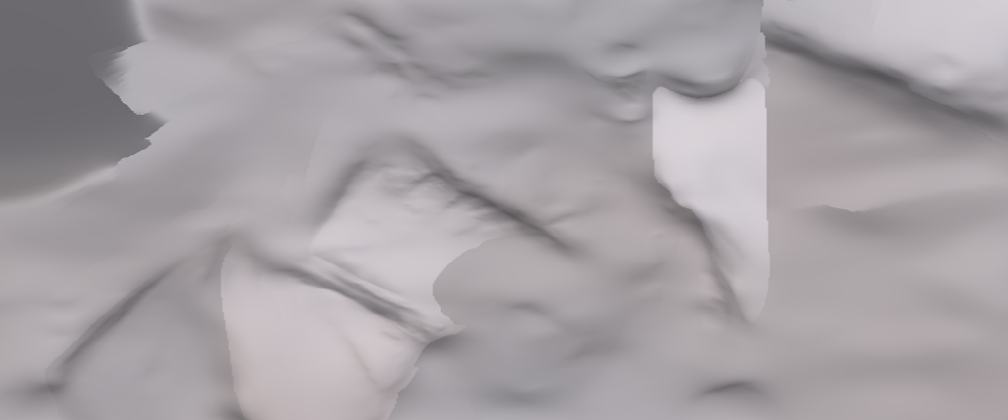}}\hfill
		\subfigure[]
		{\includegraphics[width=0.197\linewidth]{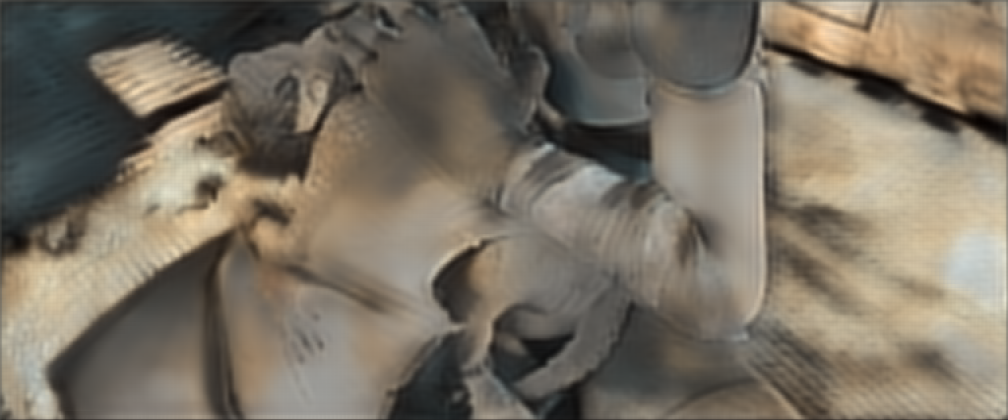}}\hfill
		\subfigure[]
		{\includegraphics[width=0.197\linewidth]{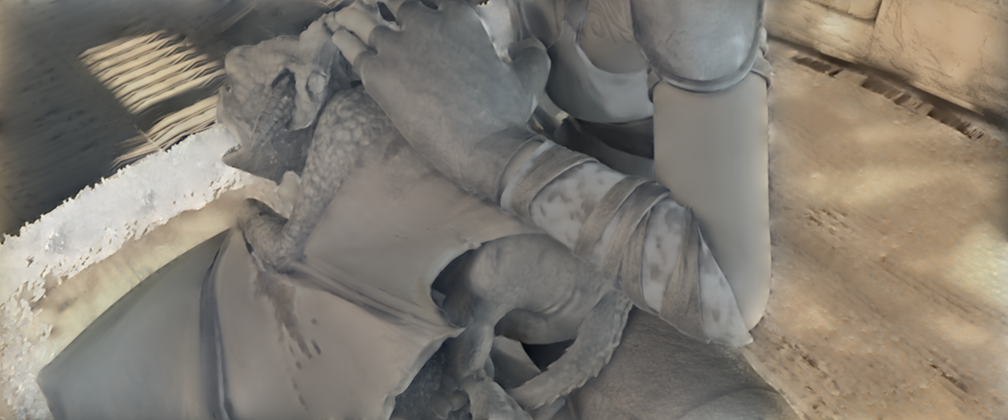}}\hfill
		\subfigure[]
		{\includegraphics[width=0.197\linewidth]{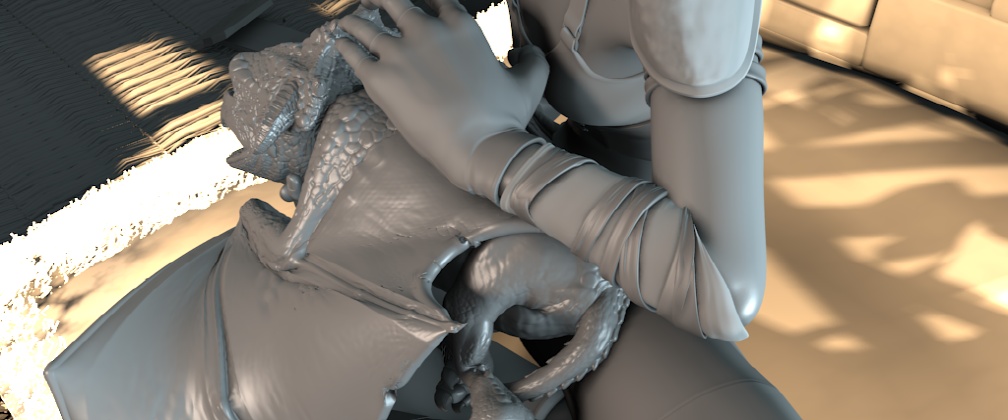}}\hfill
		\vspace{-21pt}
		\subfigure[]
		{\includegraphics[width=0.197\linewidth]{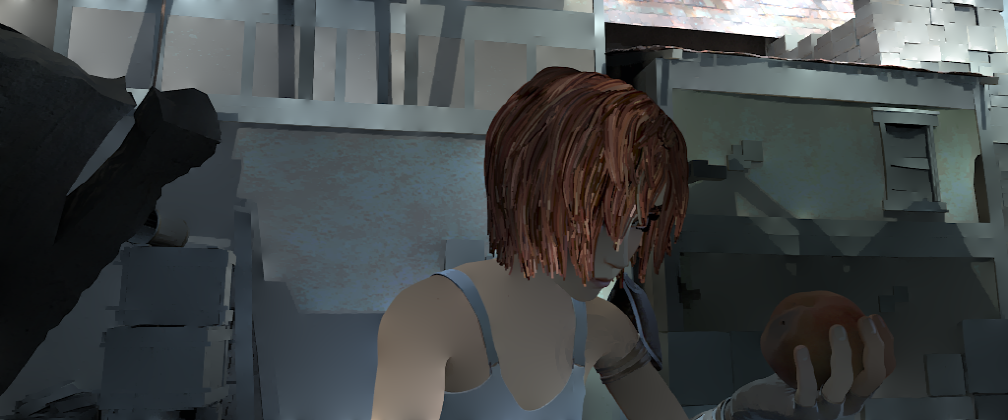}}\hfill
		\subfigure[]
		{\includegraphics[width=0.197\linewidth]{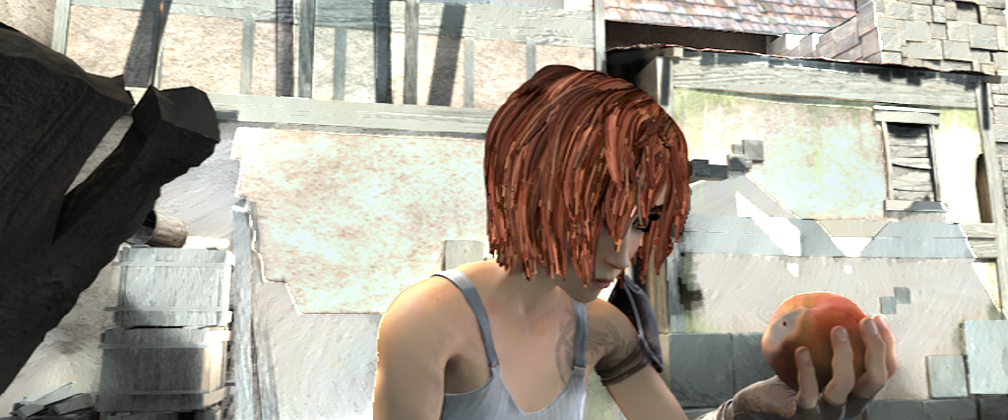}}\hfill
		\subfigure[]
		{\includegraphics[width=0.197\linewidth]{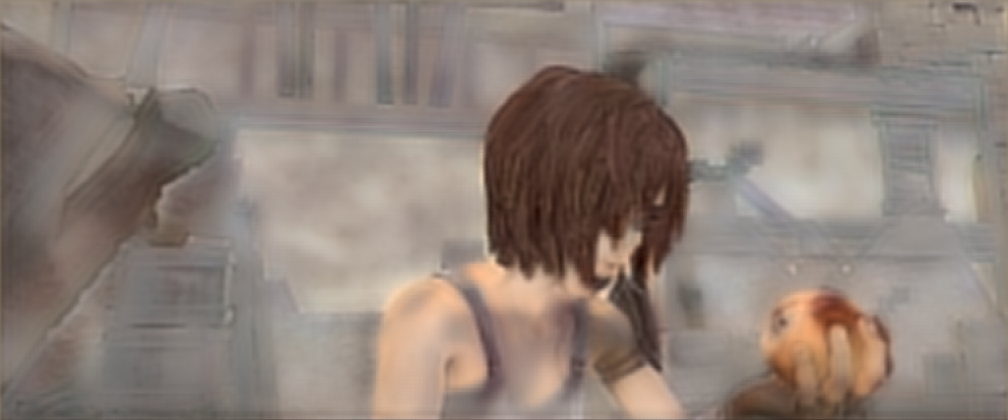}}\hfill
		\subfigure[]
		{\includegraphics[width=0.197\linewidth]{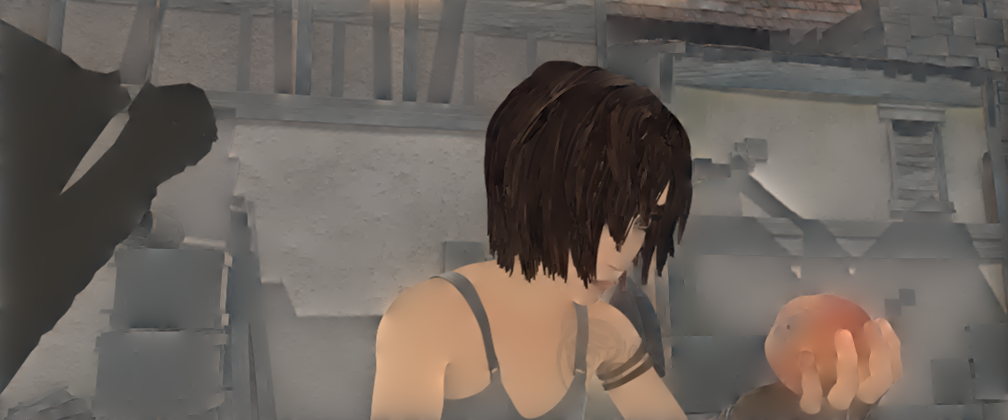}}\hfill
		\subfigure[]
		{\includegraphics[width=0.197\linewidth]{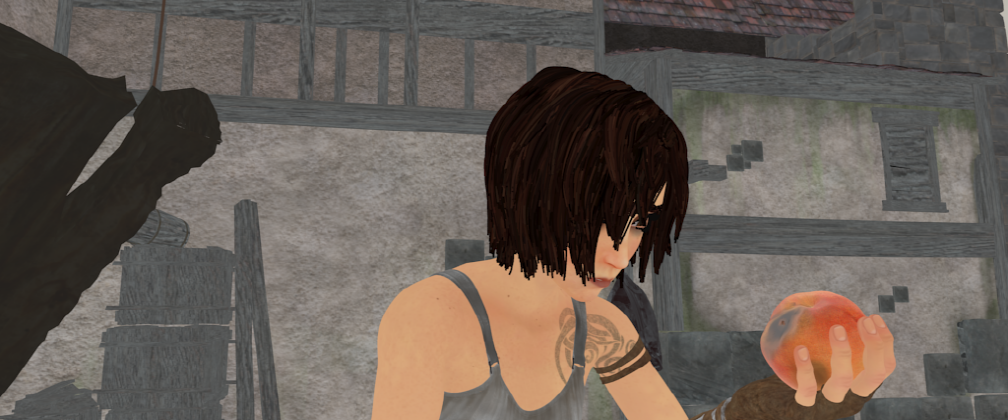}}\hfill
		\vspace{-21pt}
		\subfigure[(a)]
		{\includegraphics[width=0.197\linewidth]{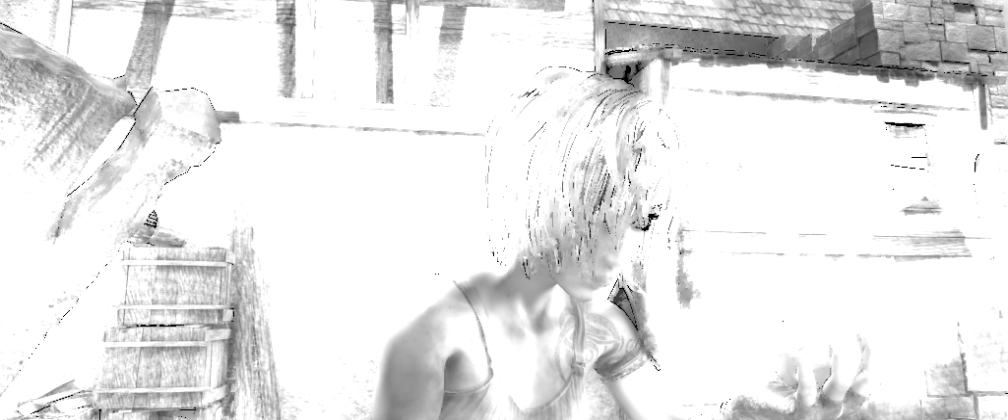}}\hfill
		\subfigure[(b)]
		{\includegraphics[width=0.197\linewidth]{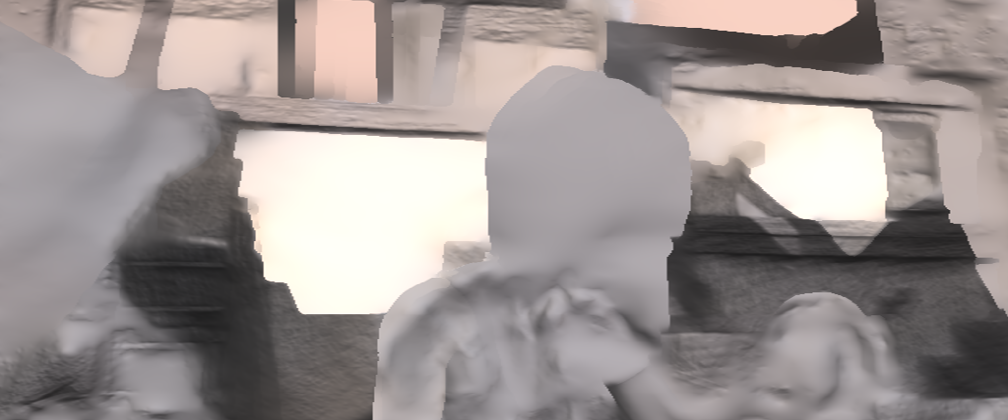}}\hfill
		\subfigure[(c)]
		{\includegraphics[width=0.197\linewidth]{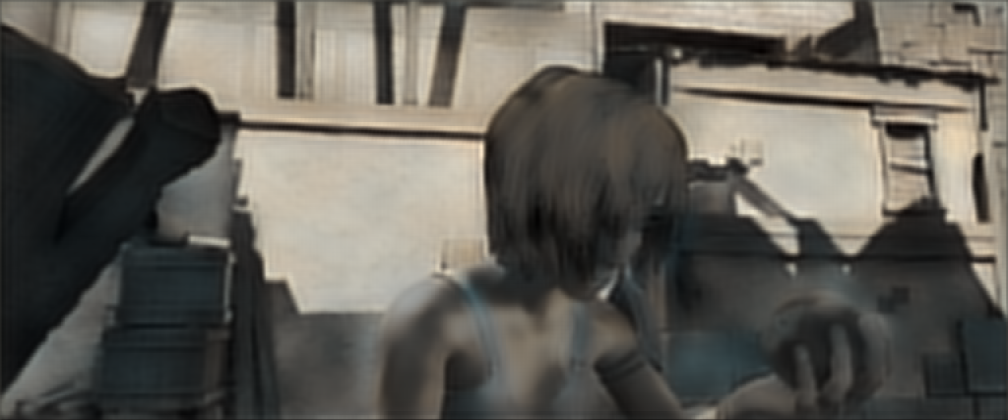}}\hfill
		\subfigure[(d)]
		{\includegraphics[width=0.197\linewidth]{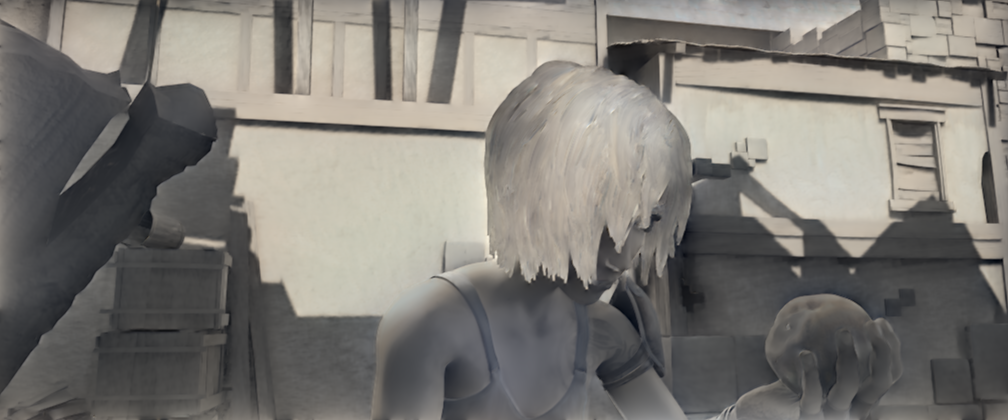}}\hfill
		\subfigure[(e)]
		{\includegraphics[width=0.197\linewidth]{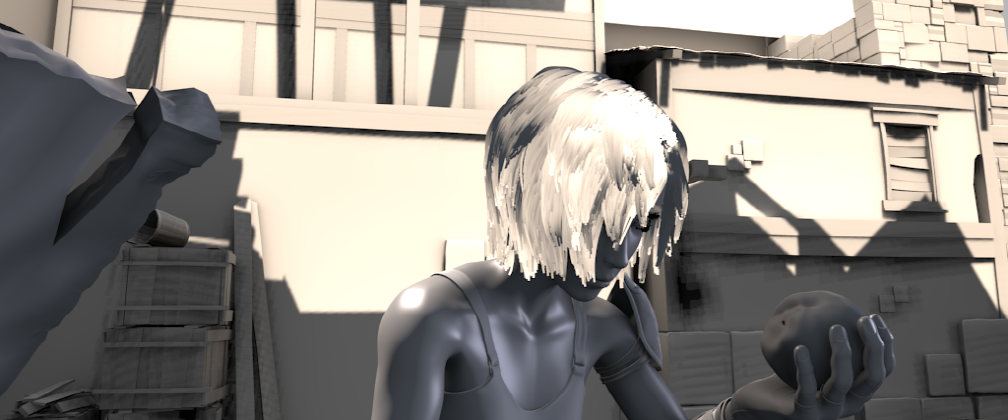}}\hfill
		\vspace{-12pt}
		\caption{Qualitative results on MPI SINTEL \cite{MPIsintel} for intrinsic decomposition of \figref{img:4}. (a) Shen \emph{et al.} \cite{Shen11}, (b) SIRFS \cite{Barron12}, (c) MSCR \cite{Narihira15}, (d) JCNF, and (e) ground truth.}\label{img:5}\vspace{-10pt}
	\end{figure}
	
	\section{Experimental Results}\label{sec:5}
	For our experiments, we implemented the JCNF model using the VLFeat MatConvNet toolbox \cite{MatConv}.
	Our code with pre-trained parameters will be made publicly available upon publication.
	
	The inputs of the global depth network were color images and the corresponding depth images at $1/16$ of the original scale.
	The inputs of each gradient network were randomly cropped patches from training images and the corresponding gradient maps.
	The patch sizes were $35\times35\times3$ for color images, $19\times19\times2$ for depth gradient maps, and $19\times19\times6$ for albedo and shading gradient maps. The reduced resolution for gradient map patches was due to boundary regions not processed by convolution \cite{Dong15}.
	For the gradient scale networks, the input patches are of size $35\times35\times9$.
	The energy function weights were set to $\{\lambda_D,\lambda_A,\lambda_S\}  = \{1,0.1,0.1\}$ by cross-validation.
	The filter weights of each network layer were initialized by drawing randomly from a Gaussian distribution with zero mean and a standard deviation of $0.001$.
	The network learning rates were set to $10^{-4}$, except for the final layer of the gradient networks where it is set to $10^{-5}$.
	
	We additionally augmented the training data by applying random transforms to it, including scalings in the range $[0.8,1.2]$, in-plane rotations in the range $[-15,15]$, translations, RGB scalings, image flips, and different gammas.
	
	In the following, we evaluated our system through comparisons to state-of-the-art depth prediction and intrinsic image decomposition methods on the MPI SINTEL \cite{MPIsintel}, NYU v2 \cite{NYUv2}, and Make3D \cite{Make3D} benchmarks. We additionally examined the performance contributions of the joint network learning (wo/jnl), the gradient scale network (wo/gsn), and the coarse-to-fine scheme (wo/ctf).
	\vspace{-5pt}
	\begin{table}[t]
		\centering
		\begin{tabular}{ >{\raggedright}m{0.27\linewidth}
				>{\centering}m{0.08\linewidth}  >{\centering}m{0.08\linewidth}
				>{\centering}m{0.08\linewidth}  >{\centering}m{0.08\linewidth}
				>{\centering}m{0.11\linewidth}  >{\centering}m{0.11\linewidth}
				>{\centering}m{0.11\linewidth}  }
			\hlinewd{0.8pt}
			\multirow{2}{*}{$\;$Methods} & \multicolumn{4}{ c }{Error}
			& \multicolumn{3}{ c }{Accuracy} \tabularnewline
			\cline{2-8}
			&rel &log$_{10}$ &rms &rms$_\mathrm{log}$ &{\scriptsize$\delta<1.25$} &{\scriptsize$\delta<1.25^2$} &{\scriptsize$\delta<1.25^3$} \tabularnewline
			\hline
			\hline
			$\;$Depth Transfer \cite{Karsch14} &0.448 &0.193 &9.242 &3.121 &0.524 &0.712 &0.735 \tabularnewline
			$\;$Depth Analogy \cite{Choi15} &0.432 &0.167 &8.421 &2.741 &0.621 &0.799 &0.812 \tabularnewline
			$\;$DCNF-FCSP{\scriptsize(NYU)} \cite{Fayao15} &0.424 &0.164 &8.112 &2.421 &0.652 &0.782 &0.824 \tabularnewline
			\hline
			$\;$\textbf{JCNF{\scriptsize(NYU)}}\cellcolor{blue!5} &\textbf{0.293}\cellcolor{blue!5} &\textbf{0.131}\cellcolor{blue!5} &\textbf{7.421}\cellcolor{blue!5} &\textbf{1.812}\cellcolor{blue!5} &\textbf{0.715}\cellcolor{blue!5} &\textbf{0.812}\cellcolor{blue!5} &\textbf{0.831}\cellcolor{blue!5} \tabularnewline
			$\;$\textbf{JCNF wo/jnl}\cellcolor{blue!5} &\textbf{0.292}\cellcolor{blue!5} &\textbf{0.138}\cellcolor{blue!5} &\textbf{7.471}\cellcolor{blue!5} &\textbf{1.973}\cellcolor{blue!5} &\textbf{0.714}\cellcolor{blue!5} &\textbf{0.783}\cellcolor{blue!5} &\textbf{0.839}\cellcolor{blue!5} \tabularnewline
			$\;$\textbf{JCNF wo/gsn}\cellcolor{blue!5} &\textbf{0.271}\cellcolor{blue!5} &\textbf{0.119}\cellcolor{blue!5} &\textbf{7.451}\cellcolor{blue!5} &\textbf{1.921}\cellcolor{blue!5} &\textbf{0.724}\cellcolor{blue!5} &\textbf{0.793}\cellcolor{blue!5} &\textbf{0.893}\cellcolor{blue!5} \tabularnewline
			$\;$\textbf{JCNF wo/ctf}\cellcolor{blue!5} &\textbf{0.252}\cellcolor{blue!5} &\textbf{0.101}\cellcolor{blue!5} &\textbf{7.233}\cellcolor{blue!5} &\textbf{1.622}\cellcolor{blue!5} &\textbf{0.729}\cellcolor{blue!5} &\textbf{0.812}\cellcolor{blue!5} &\textbf{0.878}\cellcolor{blue!5} \tabularnewline
			$\;$\textbf{JCNF}\cellcolor{blue!5} &\textbf{0.183}\cellcolor{blue!5} &\textbf{0.097}\cellcolor{blue!5} &\textbf{6.118}\cellcolor{blue!5} &\textbf{1.037}\cellcolor{blue!5} &\textbf{0.823}\cellcolor{blue!5} &\textbf{0.834}\cellcolor{blue!5} &\textbf{0.902}\cellcolor{blue!5} \tabularnewline
			\hlinewd{0.8pt}
		\end{tabular}\vspace{+3pt}
		\caption{Quantitative results on MPI SINTEL \cite{MPIsintel} for depth prediction. DCNF-FCSP {\scriptsize(NYU)} \cite{Fayao15} and JCNF{\scriptsize(NYU)} predict the depth by pre-training on NYU v2 \cite{NYUv2}.}\label{tab:2}\vspace{-15pt}
	\end{table}
	\begin{table}[t]
		\centering
		\begin{tabular}{ >{\raggedright}m{0.2\linewidth}
				>{\centering}m{0.08\linewidth}  >{\centering}m{0.08\linewidth}
				>{\centering}m{0.08\linewidth}  >{\centering}m{0.08\linewidth}
				>{\centering}m{0.08\linewidth}  >{\centering}m{0.08\linewidth}
				>{\centering}m{0.08\linewidth}  >{\centering}m{0.08\linewidth}
				>{\centering}m{0.08\linewidth} }
			\hlinewd{0.8pt}
			\multirow{2}{*}{$\;$Methods} & \multicolumn{3}{ c }{MSE}
			& \multicolumn{3}{ c }{LMSE} & \multicolumn{3}{ c }{DSSIM} \tabularnewline
			\cline{2-10}
			&albedo &shading &avg. &albedo &shading &avg. &albedo &shading &avg. \tabularnewline
			\hline
			\hline
			$\;$Retinex \cite{Grosse09} &0.053 &0.049 &0.051 &0.033 &0.028 &0.031 &0.214 &0.206 &0.210 \tabularnewline
			$\;$Li \emph{et al.} \cite{Li14} &0.042 &0.041 &0.037 &0.024 &0.031 &0.034 &0.242 &0.224 &0.194 \tabularnewline
			$\;$Shen \emph{et al.} \cite{Shen11} &0.043 &0.039 &0.048 &0.028 &0.027 &0.032 &0.221 &0.210 &0.232 \tabularnewline
			$\;$Zhao \emph{et al.} \cite{Zhao12} &0.047 &0.041 &0.031 &0.028 &0.029 &0.031 &0.210 &0.257 &0.214 \tabularnewline
			$\;$IIW \cite{Bell14} &0.041 &0.032 &0.041 &0.032  &0.031 &0.027 &0.281 &0.241 &0.284 \tabularnewline
			$\;$SIRFS \cite{Barron12} &0.042 &0.047 &0.043 &0.029 &0.026 &0.028 &0.210 &0.206 &0.208 \tabularnewline
			\hline
			$\;$Jeon \emph{et al.} \cite{Jeon14} &0.042 &0.033 &0.032 &0.021 &0.021 &0.023 &0.204 &0.181 &0.193 \tabularnewline
			$\;$Chen \emph{et al.} \cite{Chen13} &0.031 &0.028 &0.029 &0.019 &0.019 &0.019 &0.196 &0.165 &0.181 \tabularnewline
			\hline
			$\;$MSCR \cite{Narihira15} &0.020 &0.017 &0.021 &0.016 &0.011 &0.011 &0.201 &0.150 &0.176 \tabularnewline
			\hline
			$\;$\textbf{JCNF wo/jnl}\cellcolor{blue!5} &\textbf{0.012}\cellcolor{blue!5} &\textbf{0.015}\cellcolor{blue!5} &\textbf{0.016}\cellcolor{blue!5} &\textbf{0.014}\cellcolor{blue!5} &\textbf{0.010}\cellcolor{blue!5} &\textbf{0.010}\cellcolor{blue!5} &\textbf{0.149}\cellcolor{blue!5} &\textbf{0.123}\cellcolor{blue!5} &\textbf{0.141}\cellcolor{blue!5} \tabularnewline
			$\;$\textbf{JCNF wo/gsn}\cellcolor{blue!5} &\textbf{0.008}\cellcolor{blue!5} &\textbf{0.011}\cellcolor{blue!5} &\textbf{0.011}\cellcolor{blue!5} &\textbf{0.010}\cellcolor{blue!5} &\textbf{0.009}\cellcolor{blue!5} &\textbf{0.008}\cellcolor{blue!5} &\textbf{0.146}\cellcolor{blue!5} &\textbf{0.112}\cellcolor{blue!5} &\textbf{0.132}\cellcolor{blue!5} \tabularnewline
			$\;$\textbf{JCNF wo/ctf}\cellcolor{blue!5} &\textbf{0.008}\cellcolor{blue!5} &\textbf{0.012}\cellcolor{blue!5} &\textbf{0.010}\cellcolor{blue!5} &\textbf{0.009}\cellcolor{blue!5} &\textbf{0.008}\cellcolor{blue!5} &\textbf{0.008}\cellcolor{blue!5} &\textbf{0.127}\cellcolor{blue!5} &\textbf{0.110}\cellcolor{blue!5} &\textbf{0.119}\cellcolor{blue!5} \tabularnewline
			$\;$\textbf{JCNF}\cellcolor{blue!5} &\textbf{0.007}\cellcolor{blue!5} &\textbf{0.009}\cellcolor{blue!5} &\textbf{0.007}\cellcolor{blue!5} &\textbf{0.006}\cellcolor{blue!5} &\textbf{0.007}\cellcolor{blue!5} &\textbf{0.007}\cellcolor{blue!5} &\textbf{0.092}\cellcolor{blue!5} &\textbf{0.101}\cellcolor{blue!5} &\textbf{0.097}\cellcolor{blue!5} \tabularnewline
			\hlinewd{0.8pt}
		\end{tabular}\vspace{+3pt}
		\caption{Quantitative results on MPI SINTEL \cite{MPIsintel} for intrinsic decomposition using methods based on single images, RGB-D, CNNs, and our JCNF model.}\label{tab:3}\vspace{-20pt}
	\end{table}
	
	\subsection{MPI SINTEL Benchmark}\label{sec:52}
	We evaluated our JCNF model on both depth prediction and intrinsic image decomposition on the MPI SINTEL benchmark \cite{MPIsintel},
	which consists of $890$ images from $18$ scenes with $50$ frames each.
	For a fair evaluation,
	we followed the same experimental protocol as in \cite{Chen13,Narihira15},
	with their two-fold cross-validation and training/testing image splits.
	\figref{img:4} and \figref{img:5} exhibit predicted depth and intrinsic images
	from a single image, respectively.
	\tabref{tab:2} and \tabref{tab:3} are quantitative evaluations for both tasks using a variety of metrics, including average relative difference (rel), average log$_{10}$ error (log$_{10}$),
	root-mean-squared error (rms), its log version (rms$_\mathrm{log}$),
	and accuracy with thresholds $\delta=\{1.25,1.25^2,1.25^3\}$ \cite{Fayao15}.
	For quantitatively evaluating intrinsic image decomposition performance, we used
	mean-squared error (MSE), local mean-squared error (LMSE),
	and the dissimilarity version of the structural similarity index (DSSIM) \cite{Narihira15}.
	\begin{table}[t]
		\centering
		\begin{tabular}{ >{\raggedright}m{0.24\linewidth}
				>{\centering}m{0.08\linewidth}  >{\centering}m{0.08\linewidth}
				>{\centering}m{0.08\linewidth}  >{\centering}m{0.08\linewidth}
				>{\centering}m{0.11\linewidth}  >{\centering}m{0.11\linewidth}
				>{\centering}m{0.11\linewidth}  }
			\hlinewd{0.8pt}
			\multirow{2}{*}{$\;$Methods} & \multicolumn{4}{ c }{Error}
			& \multicolumn{3}{ c }{Accuracy} \tabularnewline
			\cline{2-8}
			&rel &log$_{10}$ &rms &rms$_\mathrm{log}$
			&{\scriptsize$\delta<1.25$} &{\scriptsize$\delta<1.25^2$} &{\scriptsize$\delta<1.25^3$} \tabularnewline
			\hline
			\hline
			$\;$Make3D \cite{Saxena09} &0.349 &- &1.214 &0.409 &0.447 &0.745 &0.897 \tabularnewline
			$\;$Depth Transfer \cite{Karsch14} &0.350 &0.134 &1.1 &0.378 &0.460 &0.742 &0.893 \tabularnewline
			$\;$Depth Analogy \cite{Choi15} &0.328 &0.132 &1.31 &0.392 &0.471 &0.799 &0.891 \tabularnewline
			$\;$MS-CNNs \cite{Eigen14} &0.228 &- &0.901 &0.293 &0.611 &0.873 &0.961 \tabularnewline
			$\;$DCNF-FCSP \cite{Fayao15} &0.221 &0.095 &0.760 &0.281  &0.604 &0.885 &0.974 \tabularnewline
			\hline
			$\;$\textbf{JCNF{\scriptsize(MPI)}}\cellcolor{blue!5} &\textbf{0.214}\cellcolor{blue!5} &\textbf{0.093}\cellcolor{blue!5} &\textbf{0.716}\cellcolor{blue!5} &\textbf{0.241}\cellcolor{blue!5} &\textbf{0.677}\cellcolor{blue!5} &\textbf{0.879}\cellcolor{blue!5} &\textbf{0.927}\cellcolor{blue!5} \tabularnewline
			$\;$\textbf{JCNF wo/jnl}\cellcolor{blue!5} &\textbf{0.216}\cellcolor{blue!5} &\textbf{0.101}\cellcolor{blue!5} &\textbf{0.753}\cellcolor{blue!5} &\textbf{0.241}\cellcolor{blue!5} &\textbf{0.625}\cellcolor{blue!5} &\textbf{0.896}\cellcolor{blue!5} &\textbf{0.925}\cellcolor{blue!5} \tabularnewline
			$\;$\textbf{JCNF wo/gsn}\cellcolor{blue!5} &\textbf{0.210}\cellcolor{blue!5} &\textbf{0.091}\cellcolor{blue!5} &\textbf{0.728}\cellcolor{blue!5} &\textbf{0.254}\cellcolor{blue!5} &\textbf{0.621}\cellcolor{blue!5} &\textbf{0.890}\cellcolor{blue!5} &\textbf{0.975}\cellcolor{blue!5} \tabularnewline
			$\;$\textbf{JCNF wo/ctf}\cellcolor{blue!5} &\textbf{0.208}\cellcolor{blue!5} &\textbf{0.106}\cellcolor{blue!5} &\textbf{0.708}\cellcolor{blue!5} &\textbf{0.237}\cellcolor{blue!5} &\textbf{0.681}\cellcolor{blue!5} &\textbf{0.901}\cellcolor{blue!5} &\textbf{0.972}\cellcolor{blue!5} \tabularnewline
			$\;$\textbf{JCNF}\cellcolor{blue!5} &\textbf{0.201}\cellcolor{blue!5} &\textbf{0.077}\cellcolor{blue!5} &\textbf{0.711}\cellcolor{blue!5} &\textbf{0.212}\cellcolor{blue!5} &\textbf{0.690}\cellcolor{blue!5} &\textbf{0.910}\cellcolor{blue!5} &\textbf{0.979}\cellcolor{blue!5} \tabularnewline
			\hlinewd{0.8pt}
		\end{tabular}\vspace{+3pt}
		\caption{Quantitative results on the NYU v2 dataset \cite{NYUv2} for depth prediction.}\label{tab:4}\vspace{-20pt}
	\end{table}
	\begin{figure}[t]
		\centering
		\renewcommand{\thesubfigure}{}
		\subfigure[(a)]
		{\includegraphics[width=0.163\linewidth]{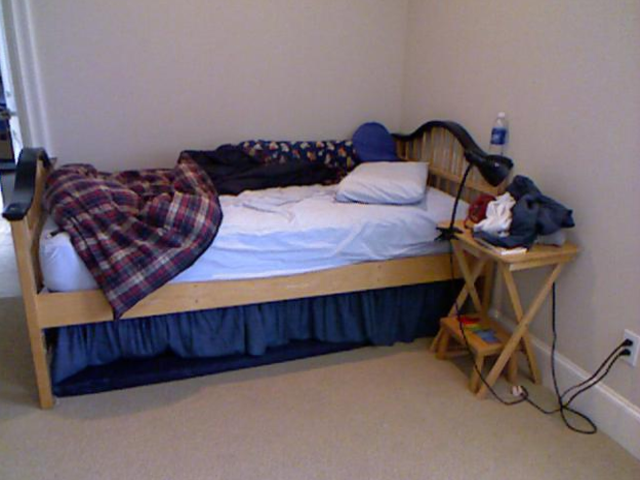}}\hfill
		\subfigure[(b)]
		{\includegraphics[width=0.163\linewidth]{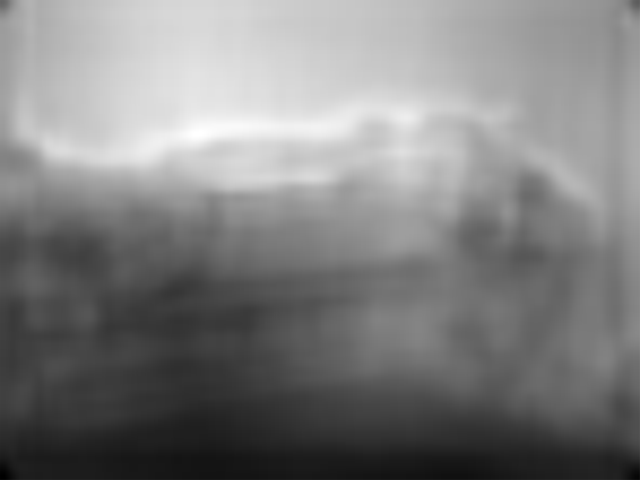}}\hfill
		\subfigure[(c)]
		{\includegraphics[width=0.163\linewidth]{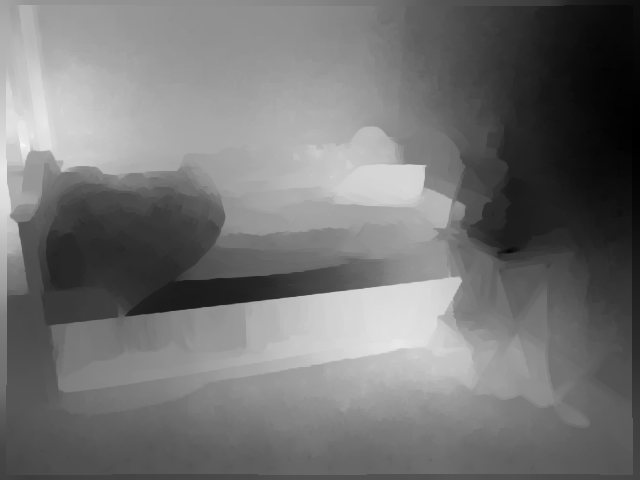}}\hfill
		\subfigure[(d)]
		{\includegraphics[width=0.163\linewidth]{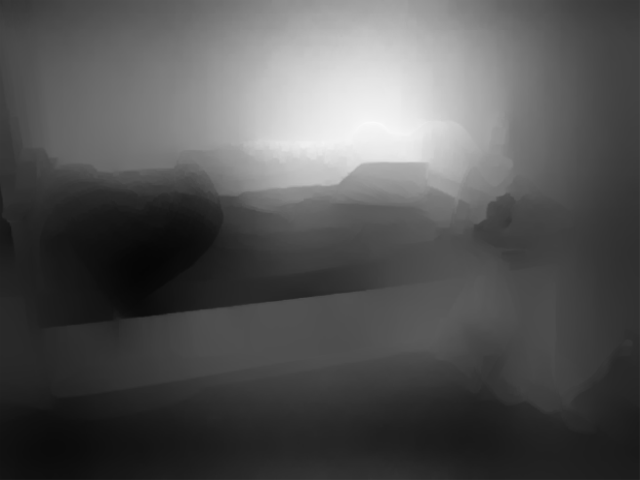}}\hfill
		\subfigure[(e)]
		{\includegraphics[width=0.163\linewidth]{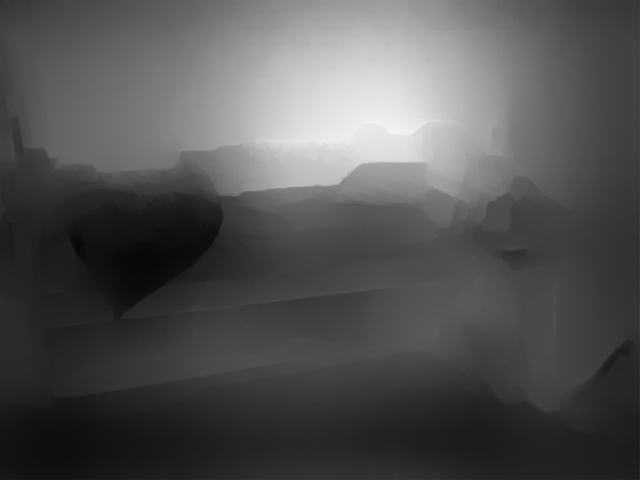}}\hfill
		\subfigure[(f)]
		{\includegraphics[width=0.163\linewidth]{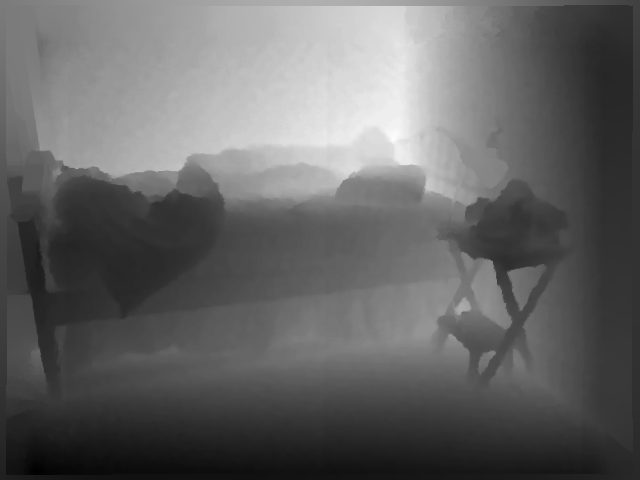}}\hfill
		\vspace{-12pt}
		\caption{Qualitative results on NYU v2 \cite{NYUv2} for depth prediction. (a) color image, (b) MS-CNNs \cite{Eigen14}, (c) DCNF-FCSP \cite{Fayao15}, (d) JCNF{\scriptsize(MPI)}, (e) JCNF, and (f) ground truth.}\label{img:6}\vspace{-10pt}
	\end{figure}
	\begin{figure}[!t]
		\centering
		\renewcommand{\thesubfigure}{}
		\subfigure[]
		{\includegraphics[width=0.163\linewidth]{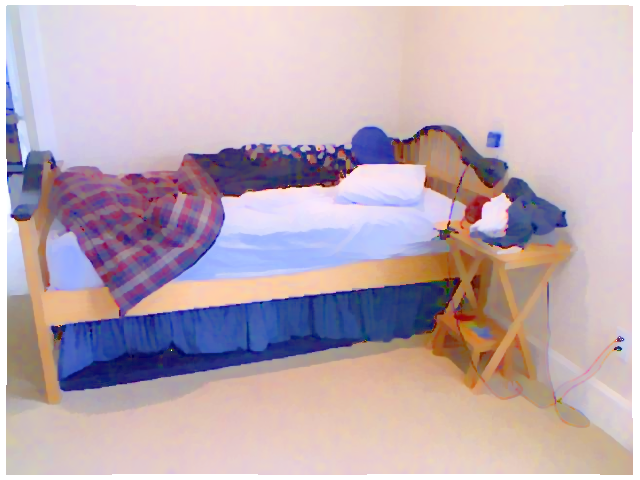}}\hfill
		\subfigure[]
		{\includegraphics[width=0.163\linewidth]{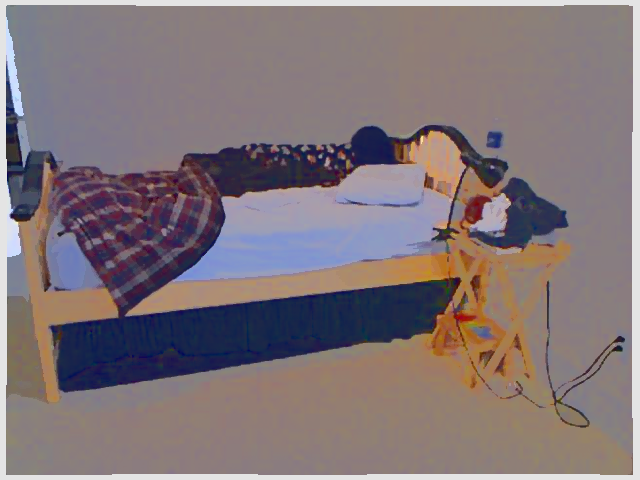}}\hfill
		\subfigure[]
		{\includegraphics[width=0.163\linewidth]{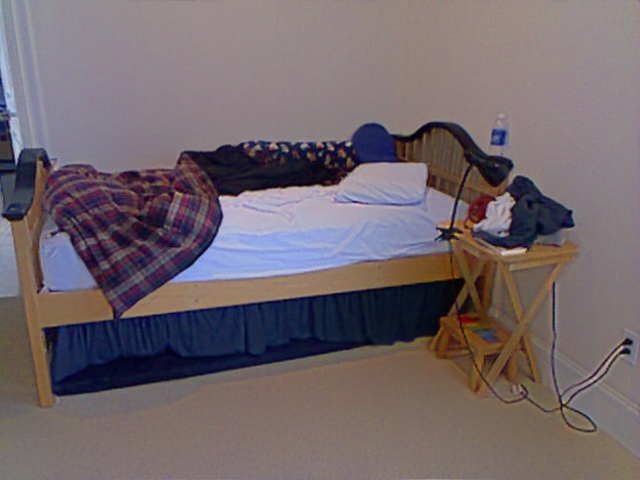}}\hfill
		\subfigure[]
		{\includegraphics[width=0.163\linewidth]{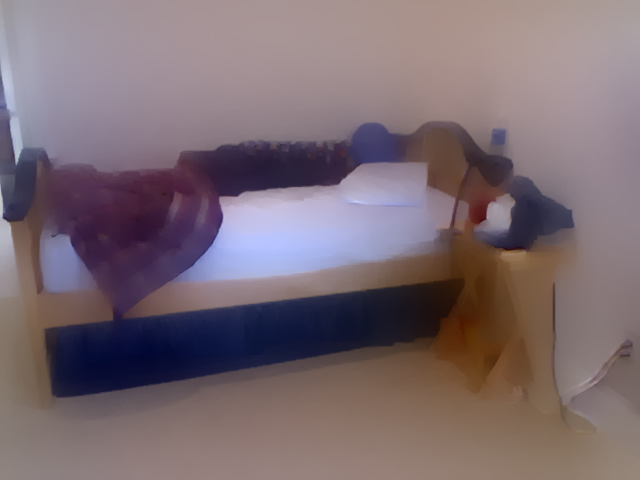}}\hfill
		\subfigure[]
		{\includegraphics[width=0.163\linewidth]{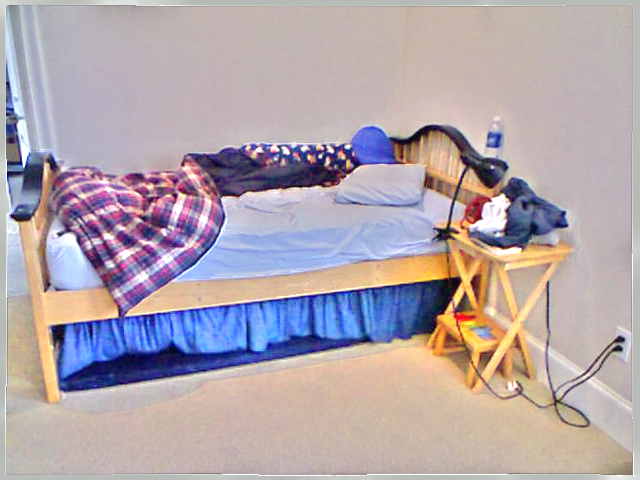}}\hfill
		\subfigure[]
		{\includegraphics[width=0.163\linewidth]{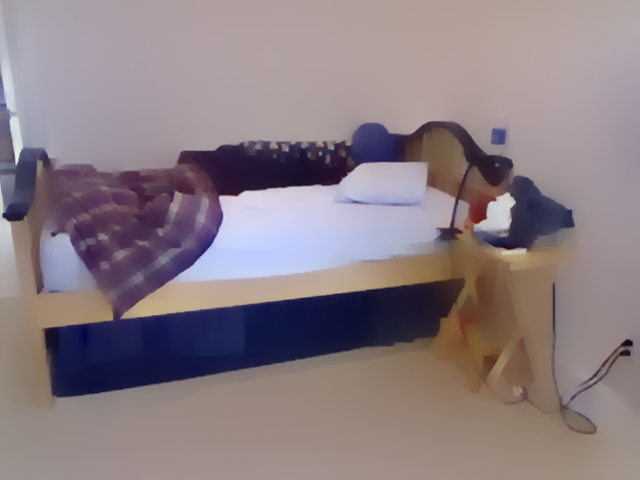}}\hfill
		\vspace{-21pt}
		\subfigure[(a)]
		{\includegraphics[width=0.163\linewidth]{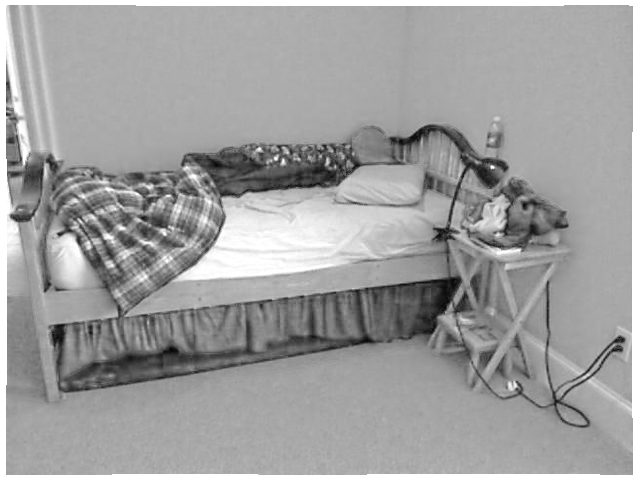}}\hfill
		\subfigure[(b)]
		{\includegraphics[width=0.163\linewidth]{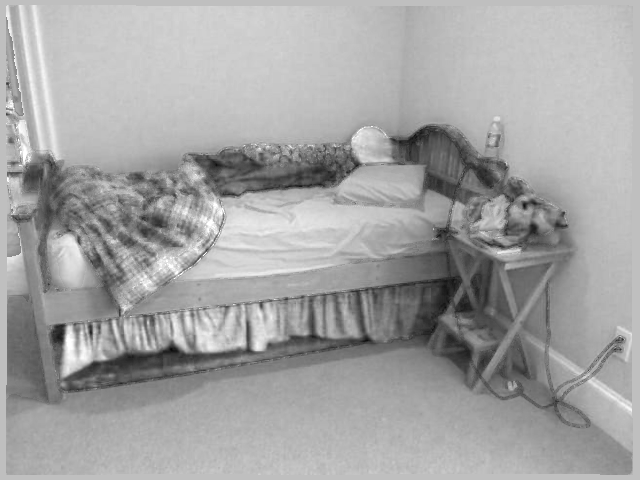}}\hfill
		\subfigure[(c)]
		{\includegraphics[width=0.163\linewidth]{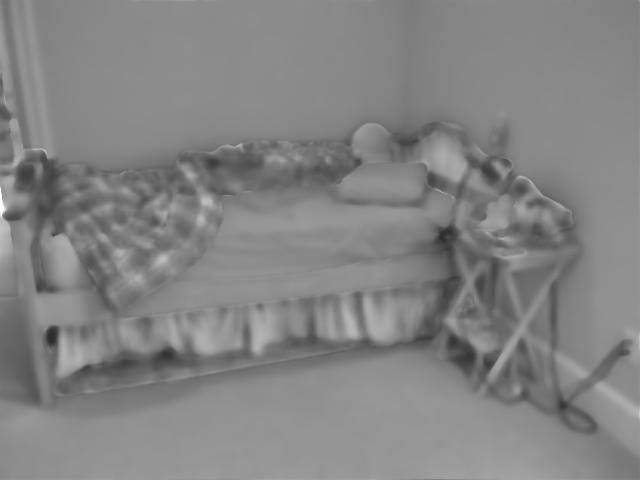}}\hfill
		\subfigure[(d)]
		{\includegraphics[width=0.163\linewidth]{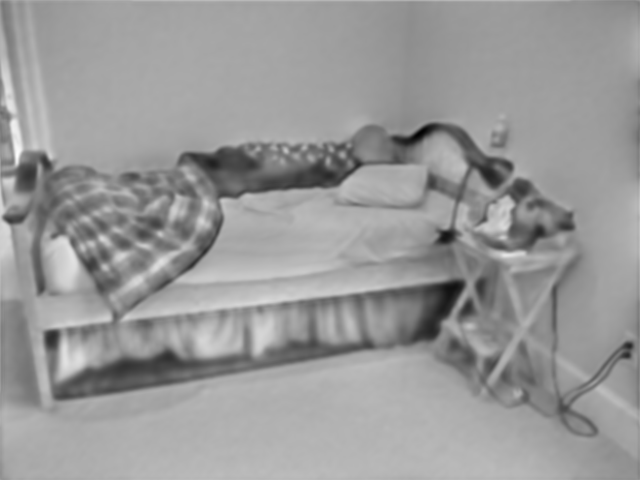}}\hfill
		\subfigure[(e)]
		{\includegraphics[width=0.163\linewidth]{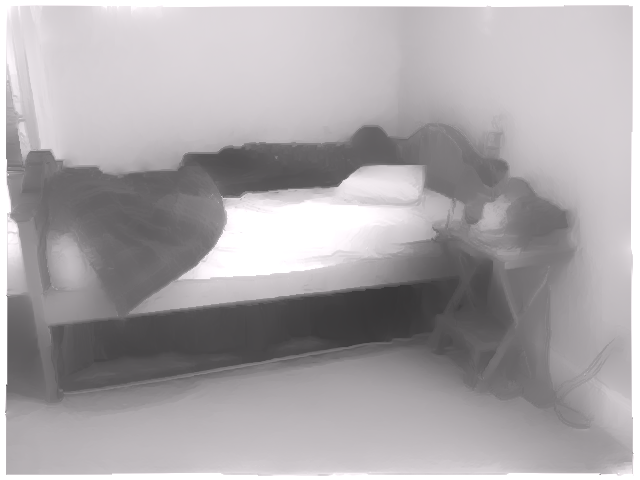}}\hfill
		\subfigure[(f)]
		{\includegraphics[width=0.163\linewidth]{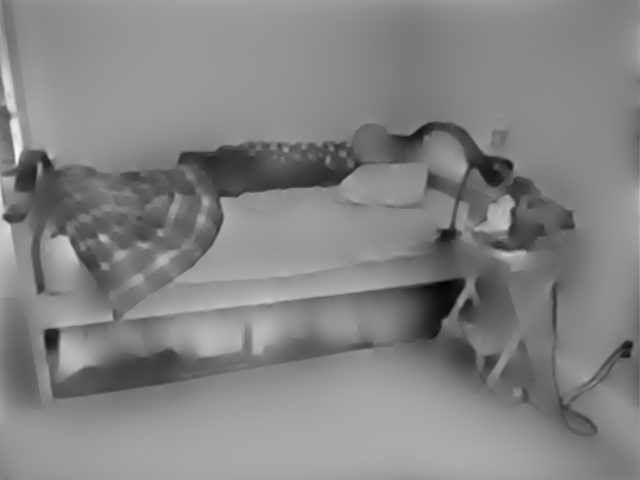}}\hfill
		\vspace{-12pt}
		\caption{Qualitative results on NYU v2 \cite{NYUv2} for intrinsic decomposition of \figref{img:6}. (a) Li \emph{et al.} \cite{Li14}, (b) IIW \cite{Bell14}, (c) Jeon \emph{et al.} \cite{Jeon14}, (d) JCNF learned using \cite{Jeon14}, (e) Chen \emph{et al.} \cite{Chen13}, and (f) JCNF learned using  \cite{Chen13}.}\label{img:7}\vspace{-10pt}
	\end{figure}
	
	For the depth prediction task, data-driven approaches (DT \cite{Karsch14} and DA \cite{Choi15}) provided limited performance due to their low learning capacity.
	CNN-based depth prediction (DCNF-FCSP \cite{Fayao15}) using a pre-trained model from NYU v2 \cite{NYUv2} showed better performance, but is restricted by depth ambiguity problems.
	Our JCNF model achieved the best results both quantitatively and qualitatively, whether pre-trained using MPI SINTEL or NYU v2 datasets.
	Furthermore, it is shown that omitting the gradient scale network, coarse-to-fine processing, or joint learning significantly reduced depth prediction performances.
	
	In intrinsic image decomposition, existing single-image based methods \cite{Grosse09,Li14,Shen11,Zhao12,Bell14} produced the lowest quality results as they do not benefit from any additional information. RGB-D based methods \cite{Chen13,Barron13,Jeon14} performed better with measured depth as input.
	CNN-based intrinsic decomposition \cite{Narihira15} surpassed RGB-D based techniques even without having depth as an input, but its results exhibit
	some blur, likely due to ambiguity from limited training datasets. Thanks to its gradient domain learning and leverage of estimated depth information, our JCNF model provides more accurate and edge-preserved results, with the best qualitative and quantitative performance.
	\vspace{-5pt}
	\begin{table}[t]
		\centering
		\begin{tabular}{ >{\raggedright}m{0.24\linewidth}
				>{\centering}m{0.08\linewidth}  >{\centering}m{0.08\linewidth}
				>{\centering}m{0.08\linewidth}  >{\centering}m{0.08\linewidth}
				>{\centering}m{0.08\linewidth}  >{\centering}m{0.08\linewidth}
				>{\centering}m{0.08\linewidth}  >{\centering}m{0.08\linewidth} }
			\hlinewd{0.8pt}
			\multirow{2}{*}{$\;$Methods} & \multicolumn{4}{ c }{Error ($\mathrm{C}1$)}
			& \multicolumn{4}{ c }{Error ($\mathrm{C}2$)} \tabularnewline
			\cline{2-9}
			&rel &log$_{10}$ &rms &rms$_\mathrm{log}$
			&rel &log$_{10}$ &rms &rms$_\mathrm{log}$ \tabularnewline
			\hline
			\hline
			$\;$Make3D \cite{Saxena09} &0.412 &0.165 &11.1 &0.451 &0.407 &0.155 &16.1 &0.486 \tabularnewline
			$\;$Depth Transfer \cite{Karsch14} &0.355 &0.127 &9.20 &0.421 &0.438 &0.161 &14.81 &0.461 \tabularnewline
			$\;$Depth Analogy \cite{Choi15} &0.371 &0.121 &8.11 &0.381 &0.410 &0.144 &14.52 &0.479 \tabularnewline
			$\;$DCNF-FCSP \cite{Fayao15} &0.331 &0.119 &8.60 &0.392 &0.307 &0.125 &12.89 &0.412 \tabularnewline
			\hline
			$\;$\textbf{JCNF{\scriptsize(MPI)}}\cellcolor{blue!5} &\textbf{0.273}\cellcolor{blue!5} &\textbf{0.110}\cellcolor{blue!5} &\textbf{7.70}\cellcolor{blue!5} &\textbf{0.351}\cellcolor{blue!5} &\textbf{0.263}\cellcolor{blue!5} &\textbf{0.117}\cellcolor{blue!5} &\textbf{8.62}\cellcolor{blue!5} &\textbf{0.347}\cellcolor{blue!5} \tabularnewline
			$\;$\textbf{JCNF{\scriptsize(NYU)}}\cellcolor{blue!5} &\textbf{0.274}\cellcolor{blue!5} &\textbf{0.097}\cellcolor{blue!5} &\textbf{7.22}\cellcolor{blue!5} &\textbf{0.352}\cellcolor{blue!5} &\textbf{0.287}\cellcolor{blue!5} &\textbf{0.127}\cellcolor{blue!5} &\textbf{8.22}\cellcolor{blue!5} &\textbf{0.341}\cellcolor{blue!5} \tabularnewline
			$\;$\textbf{JCNF}\cellcolor{blue!5} &\textbf{0.262}\cellcolor{blue!5} &\textbf{0.092}\cellcolor{blue!5} &\textbf{6.61}\cellcolor{blue!5} &\textbf{0.321}\cellcolor{blue!5} &\textbf{0.243}\cellcolor{blue!5} &\textbf{0.091}\cellcolor{blue!5} &\textbf{6.34}\cellcolor{blue!5} &\textbf{0.302}\cellcolor{blue!5} \tabularnewline
			\hlinewd{0.8pt}
		\end{tabular}\vspace{+3pt}
		\caption{Quantitative results on the Make3D dataset \cite{Make3D} for depth prediction.}\label{tab:5}\vspace{-20pt}
	\end{table}

	\subsection{NYU v2 RGB-D Benchmark}\label{sec:53}
	For further evaluation, we obtained a set of RGB, depth, and intrinsic images by applying RGB-D based intrinsic image decomposition methods \cite{Chen13,Jeon14} on the NYU v2 RGB-D database \cite{NYUv2}. Of its $1449$ RGB-D images of indoor scenes, we used $795$ for training and $654$ for testing, which is the standard training/testing split for the dataset.
	
	For depth prediction, comparisons are made to the ground truth depth in \figref{img:6} and \tabref{tab:3} using the same experimental settings as in \cite{Fayao15}. The state-of-the-art CNN-based methods \cite{Eigen14,Fayao15}
	clearly outperformed other previous methods. The performance of our JCNF model was even higher, with pre-training on either MPI SINTEL or NYU v2. Our depth prediction network is similar to \cite{Eigen14}, but it additionally predicts depth gradients and leverages intrinsic image estimates to elevate performance.
	
	In the intrinsic image decomposition results of \figref{img:7}, the RGB-D based methods of \cite{Chen13,Jeon14} are used as ground truth for training. It is seen that our JCNF more closely resembles that assumed ground truth than single-image based techniques \cite{Li14,Bell14}.
	\vspace{-5pt}
	
	\subsection{Make3D RGB-D Benchmark}\label{sec:54}
\begin{figure}[t]
		\centering
		\renewcommand{\thesubfigure}{}
		\subfigure[(a)]
		{\includegraphics[width=0.14\linewidth]{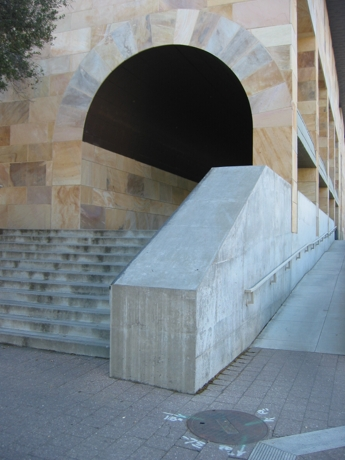}}\hfill
		\subfigure[(b)]
		{\includegraphics[width=0.14\linewidth]{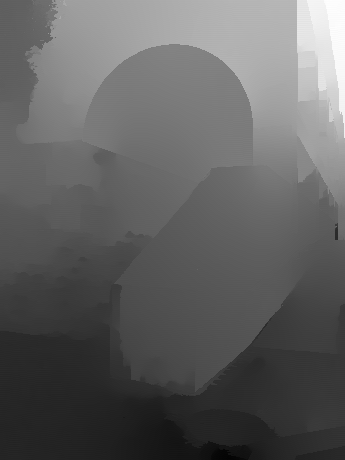}}\hfill
		\subfigure[(c)]
		{\includegraphics[width=0.14\linewidth]{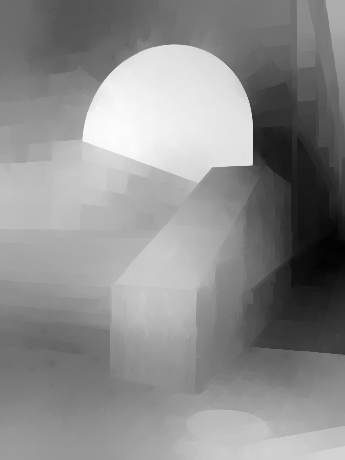}}\hfill
		\subfigure[(d)]
		{\includegraphics[width=0.14\linewidth]{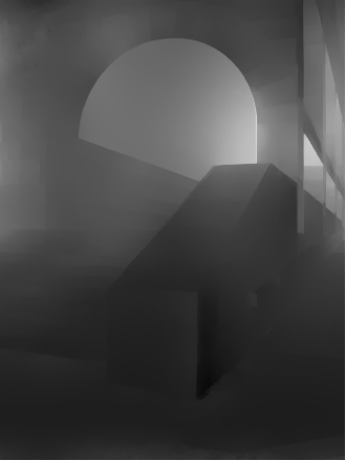}}\hfill
		\subfigure[(e)]
		{\includegraphics[width=0.14\linewidth]{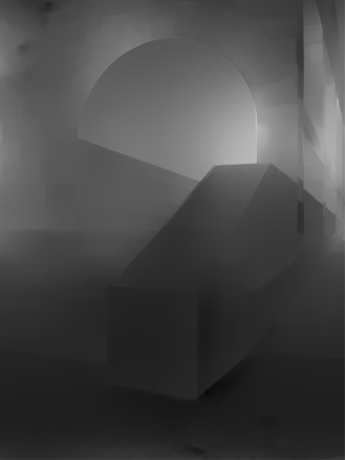}}\hfill
		\subfigure[(f)]
		{\includegraphics[width=0.14\linewidth]{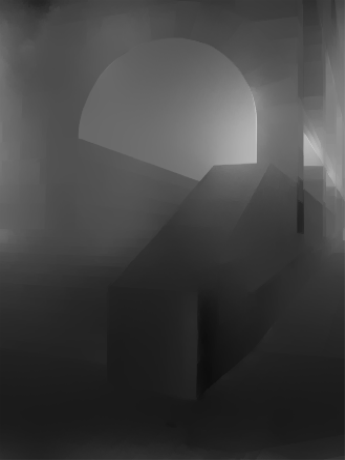}}\hfill
		\subfigure[(g)]
		{\includegraphics[width=0.14\linewidth]{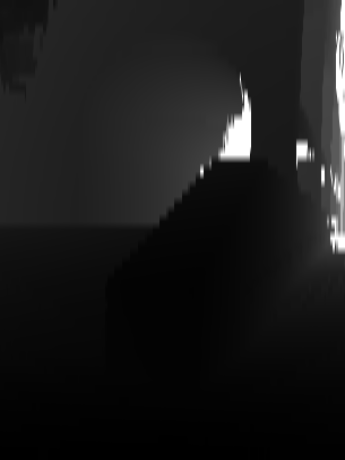}}\hfill
		\vspace{-12pt}
		\caption{Qualitative results on Make3D \cite{NYUv2} for depth prediction. (a) color image, (b) DA \cite{Choi15}, (c) DCNF-FCSP \cite{Fayao15}, (d) JCNF{\scriptsize(MPI)}, (e) JCNF{\scriptsize(NYU)}, (f) JCNF, and (g) ground truth.}\label{img:8}\vspace{-10pt}
	\end{figure}

\begin{figure}[!t]
		\centering
		\renewcommand{\thesubfigure}{}
		\subfigure[]
		{\includegraphics[width=0.14\linewidth]{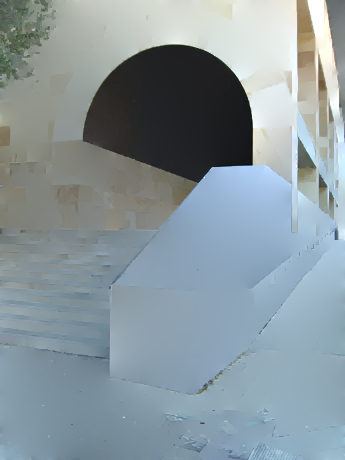}}\hfill
		\subfigure[]
		{\includegraphics[width=0.14\linewidth]{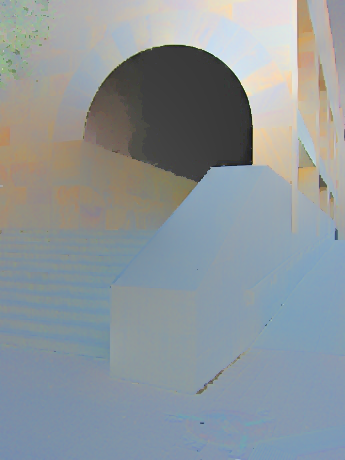}}\hfill
		\subfigure[]
		{\includegraphics[width=0.14\linewidth]{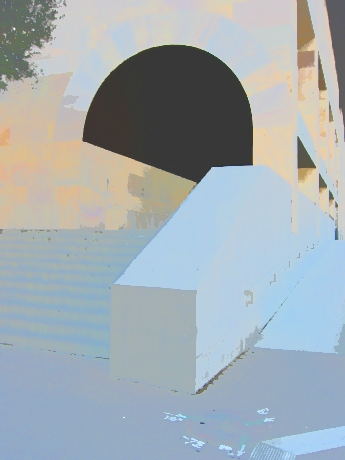}}\hfill
		\subfigure[]
		{\includegraphics[width=0.14\linewidth]{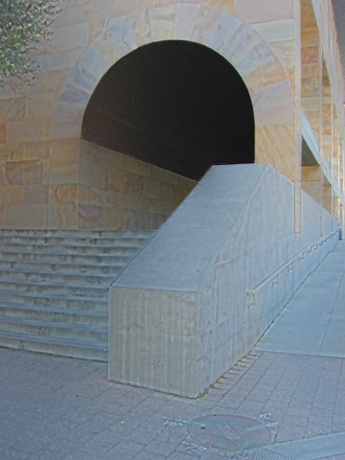}}\hfill
		\subfigure[]
		{\includegraphics[width=0.14\linewidth]{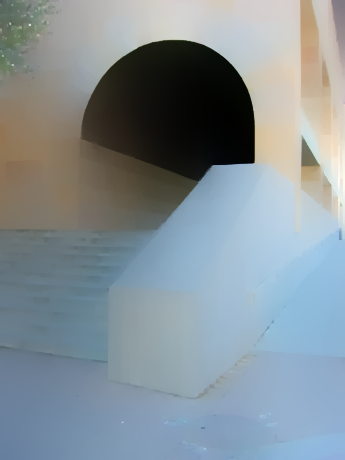}}\hfill
		\subfigure[]
		{\includegraphics[width=0.14\linewidth]{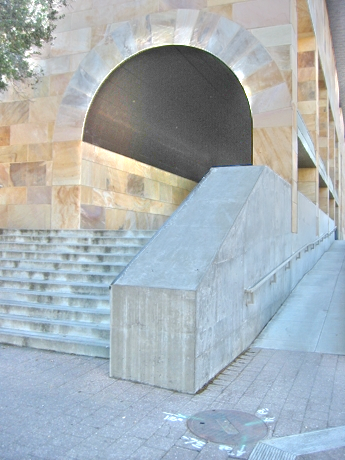}}\hfill
		\subfigure[]
		{\includegraphics[width=0.14\linewidth]{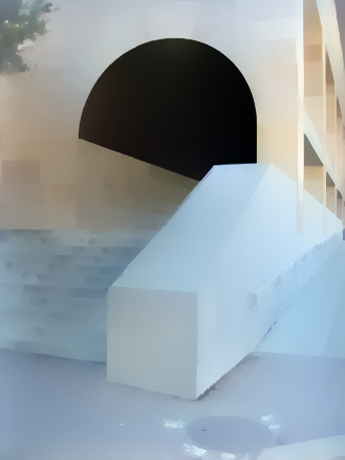}}\hfill
		\vspace{-21pt}
		\subfigure[(a)]
		{\includegraphics[width=0.14\linewidth]{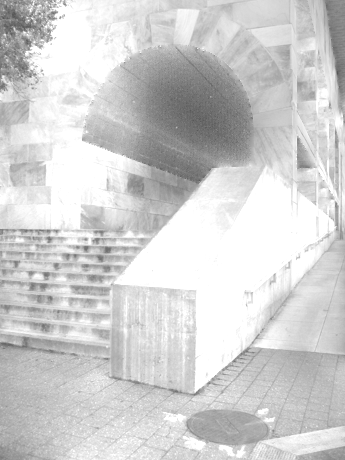}}\hfill
		\subfigure[(b)]
		{\includegraphics[width=0.14\linewidth]{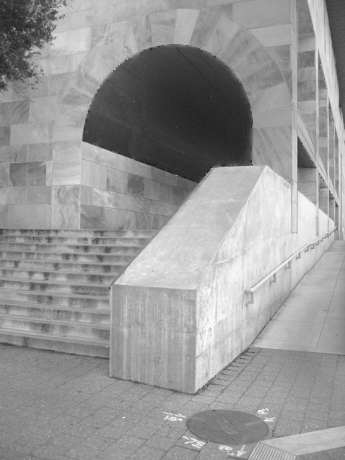}}\hfill
		\subfigure[(c)]
		{\includegraphics[width=0.14\linewidth]{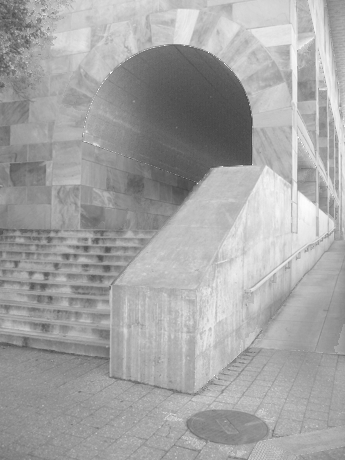}}\hfill
		\subfigure[(d)]
		{\includegraphics[width=0.14\linewidth]{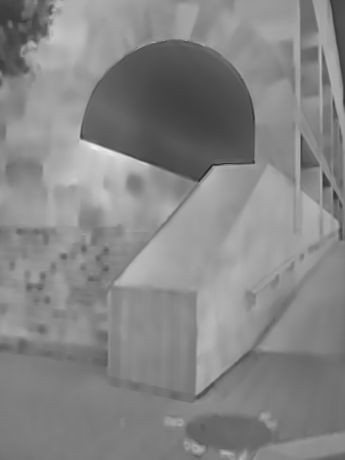}}\hfill
		\subfigure[(e)]
		{\includegraphics[width=0.14\linewidth]{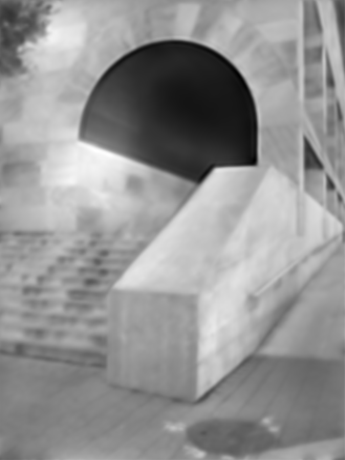}}\hfill
		\subfigure[(f)]
		{\includegraphics[width=0.14\linewidth]{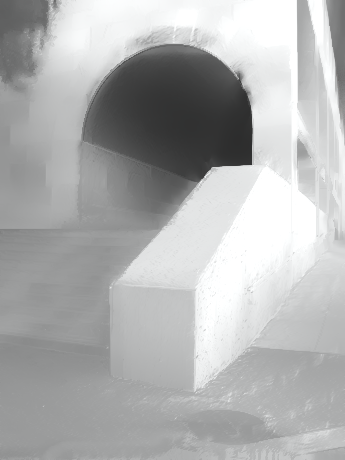}}\hfill
		\subfigure[(g)]
		{\includegraphics[width=0.14\linewidth]{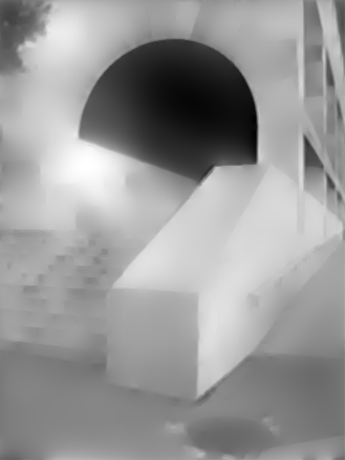}}\hfill
		\vspace{-12pt}
		\caption{Qualitative results on Make3D \cite{Make3D} for intrinsic decomposition of \figref{img:8}. (a) Li \emph{et al.} \cite{Li14}, (b) Zhao \emph{et al.} \cite{Zhao12}, (c) IIW \cite{Bell14}, (d) Jeon \emph{et al.} \cite{Jeon14}, (e) JCNF learned using \cite{Jeon14}, (f) Chen \emph{et al.} \cite{Chen13}, and (g) JCNF learned using \cite{Chen13}.}\label{img:9}\vspace{-10pt}
	\end{figure}

	We also evaluated our JCNF model on the Make3D dataset \cite{Make3D}, which contains $534$ images depicting outdoor scenes (with $400$ used for training and $134$ for testing). To account for a limitation of this dataset \cite{Saxena09,Liu14,Fayao15}, we calculate depth errors in two ways \cite{Liu14,Fayao15}: on only regions with ground truth depth less than $70$ meters (denoted by $\mathrm{C}1$), and over the entire image ($\mathrm{C}2$). From the depth prediction results in \figref{img:8} and \tabref{tab:5}, our JCNF model is found to yield the highest accuracy, even when pretrained on MPI SINTEL \cite{MPIsintel} or NYU v2 \cite{NYUv2} (\emph{i.e.}, JCNF{\scriptsize(MPI)} and JCNF{\scriptsize(NYU)}).
	For the intrinsic image decomposition results given in \figref{img:9}, JCNF also outperforms the comparison techniques. \vspace{-5pt}

	\section{Conclusion}\label{sec:6}
	\vspace{-5pt}
	We presented Joint Convolutional Neural Fields (JCNF) for jointly predicting depth, albedo and shading maps from a single input image. Its high performance can be attributed to its sharing network architecture, its gradient domain inference, and the incorporation of gradient scale network. It is shown through extensive experimentation that synergistically solving for these physical scene properties through the JCNF leads to state-of-the-art results in both single-image depth prediction and intrinsic image decomposition. In furture work, JCNF can potentially benefit shape refinement and image relighting from a single image. \vspace{-5pt}
	
	\bibliographystyle{splncs}
	\bibliography{egbib}
\end{document}